\documentclass[10pt,journal,compsoc]{IEEEtran}

\usepackage{hyperref}       
\usepackage{url}            
\usepackage{booktabs}       
\usepackage{amsfonts}       
\usepackage{graphicx}
\usepackage{subfigure}
\usepackage{amsmath}         
\usepackage{soul,xcolor}
\usepackage{multirow}
\usepackage{booktabs}
\usepackage{color, colortbl}
\definecolor{Gray}{gray}{0.9}
\usepackage{algorithm}
\usepackage{algpseudocode}
\usepackage{amsmath}
\usepackage{amssymb}
\usepackage{amsthm}
\usepackage{snapshot}

\DeclareMathOperator*{\argmin}{arg\,min}

\newtheorem{reproducing kernel}[propositionCounter]{Proposition}
\newtheorem{bandpass reconstruction}[theoremCounter]{Theorem}

\newtheorem{extended Bochner}[theoremCounter]{Theorem}
\usepackage{color}
\usepackage{dsfont}
\usepackage{multirow}
\usepackage{tabularx}

\newcommand{\comment}[1]{}

\newcommand{\bpi}{\boldsymbol{\pi}}
\newcommand{\tbpi}{\tilde{\boldsymbol{\pi}}}

\newcommand{\bP}{\mathbf{P}}
\newcommand{\tbP}{\tilde{\mathbf{P}}}
\newcommand{\bhP}{\mathbf{\hat{P}}}

\newcommand{\bdelta}{\boldsymbol{\delta}}

\newcommand{\bF}{\mathbf{F}}

\newcommand{\btheta}{{\boldsymbol{\theta}}}
\newcommand{\bhtheta}{{\boldsymbol{\hat{\theta}}}}
\newcommand{\bPhi}{{\boldsymbol{\Phi}}}
\newcommand{\bhPhi}{{\boldsymbol{\hat{\Phi}}}}

\newcommand{\tbx}[0]{\tilde{\mathbf{x}}}
\newcommand{\bx}[0]{\mathbf{x}}
\newcommand{\bz}[0]{\mathbf{z}}

\newcommand{\by}[0]{\mathbf{y}}

\newcommand{\bPsi}{\mathbf{\Psi}}

\newcommand{\bA}{\mathbf{A}}
\newcommand{\bB}{\mathbf{B}}

\DeclareMathOperator*{\argmax}{arg\,max}

\DeclareMathOperator{\argminS}{arg\,min}

\newcommand{\fig}[1]{Fig.~\ref{fig:#1}}

\newcommand{\tbl}[1]{Table~\ref{tbl:#1}}

\newcommand{\secref}[1]{Section~\ref{sec:#1}}

\newcommand{\MCTS}{\text{MCTS}}
\newcommand{\PSNR}{\text{PSNR}}
\newcommand{\sel}{\mathds{1}}

\definecolor{orange}{rgb}{1,0.5,0}
\definecolor{blue}{rgb}{0,0,0.6}

\definecolor{color1}{RGB}{0,199,1}
\definecolor{color2}{RGB}{224,43,28}

\newcommand{\eq}[1]{Eq.~\eqref{eq:#1}}

\definecolor{purple}{rgb}{0.5, 0.0, 0.8}
\definecolor{lblue}{rgb}{0.0, 0.5, 1.0}

\makeatletter
\let\amstexbig\big
\def\newbig#1{%
  \ifx#1|%
    \expandafter\@firstoftwo
  \else
    \expandafter\@secondoftwo
  \fi
  {\big@bar}%
  {\amstexbig{#1}}%
}
\AtBeginDocument{\let\big\newbig}
\def\big@bar{\bBigg@{1.1}|}
\makeatother

\newcommand{\TO}{\textbf{to} }

\newcolumntype{L}[1]{>{\hsize=#1\hsize\raggedright\arraybackslash}X}%
\newcolumntype{R}[1]{>{\hsize=#1\hsize\raggedleft\arraybackslash}X}%
\newcolumntype{C}[1]{>{\hsize=#1\hsize\centering\arraybackslash}X}%

\begin{document}
\setstcolor{red}
\title{Self-Supervised Deep Active Accelerated MRI}

 \author{Kyong Hwan~Jin,
         Michael~Unser,~\IEEEmembership{Fellow,~IEEE,}
and~Kwang Moo~Yi,~\IEEEmembership{Member,~IEEE}
\IEEEcompsocitemizethanks{\IEEEcompsocthanksitem K. H. Jin and M. Unser are with
Biomedical Imaging Group, EPFL, Lausanne, VD, Switzerland.\protect\\
E-mail: kyonghwan.jin@gmail.com, michael.unser@epfl.ch 
\IEEEcompsocthanksitem K. M. Yi is  with Visual Computing Group, University of
Victoria, BC, Canada.
E-mail: kyi@uvic.ca
}
}


 \markboth{Journal of \LaTeX\ Class Files,~Vol.~xx, No.~x, xx~xx}%
  {Shell \MakeLowercase{\textit{et al.}}: Bare Demo of IEEEtran.cls for Computer Society    Journals}


\IEEEtitleabstractindextext{
\begin{abstract}
  We propose to simultaneously learn to sample and reconstruct magnetic
  resonance images (MRI) to maximize the reconstruction quality given a
  limited sample budget, in a self-supervised setup. Unlike existing deep
  methods that focus only on reconstructing given data, thus being passive, we
  go beyond the current state of the art by considering both the data
  acquisition and the reconstruction process within a single deep-learning
  framework. As our network learns to acquire data, the network is active
  in nature.
  In order to do so, we simultaneously train two neural networks, one dedicated
  to reconstruction and the other to progressive sampling, each with an
  automatically generated supervision signal that links them together.
  The two supervision signals are created through Monte Carlo tree search
  (MCTS). MCTS returns a better sampling pattern than what the current sampling
  network can give and, thus, a better final reconstruction.  The sampling
  network is trained to mimic the MCTS results using the previous sampling
  network, thus being enhanced.  The reconstruction network is trained to give
  the highest reconstruction quality, given the MCTS sampling pattern.  Through
  this framework, we are able to train the two networks without providing any
  direct supervision on sampling.
\end{abstract}

}

\maketitle
\IEEEdisplaynontitleabstractindextext
\IEEEpeerreviewmaketitle
\section{Introduction}

\IEEEPARstart{M}{agnetic} resonance imaging (MRI) is an important noninvasive
way to investigate the human body, for example to diagnose cancer. To
obtain data, one performs scans in the frequency domain which is referred to
as \emph{k-space} in the biomedical-imaging
literature~\cite{lustig2007sparse}. Once this k-space data are acquired, they are 
then converted into the spatial domain, which provides images that physicians
  can use for diagnosis. When performing these scans, it is important
that one makes the most out of each scan, as patients can remain still and
hold their breadth for only a limited amount of time.

\def \teaserwidth {0.8}
\begin{figure}
\centering
%
 \includegraphics[trim = 60mm 15mm 70mm 37mm,clip=true,width=1\linewidth]{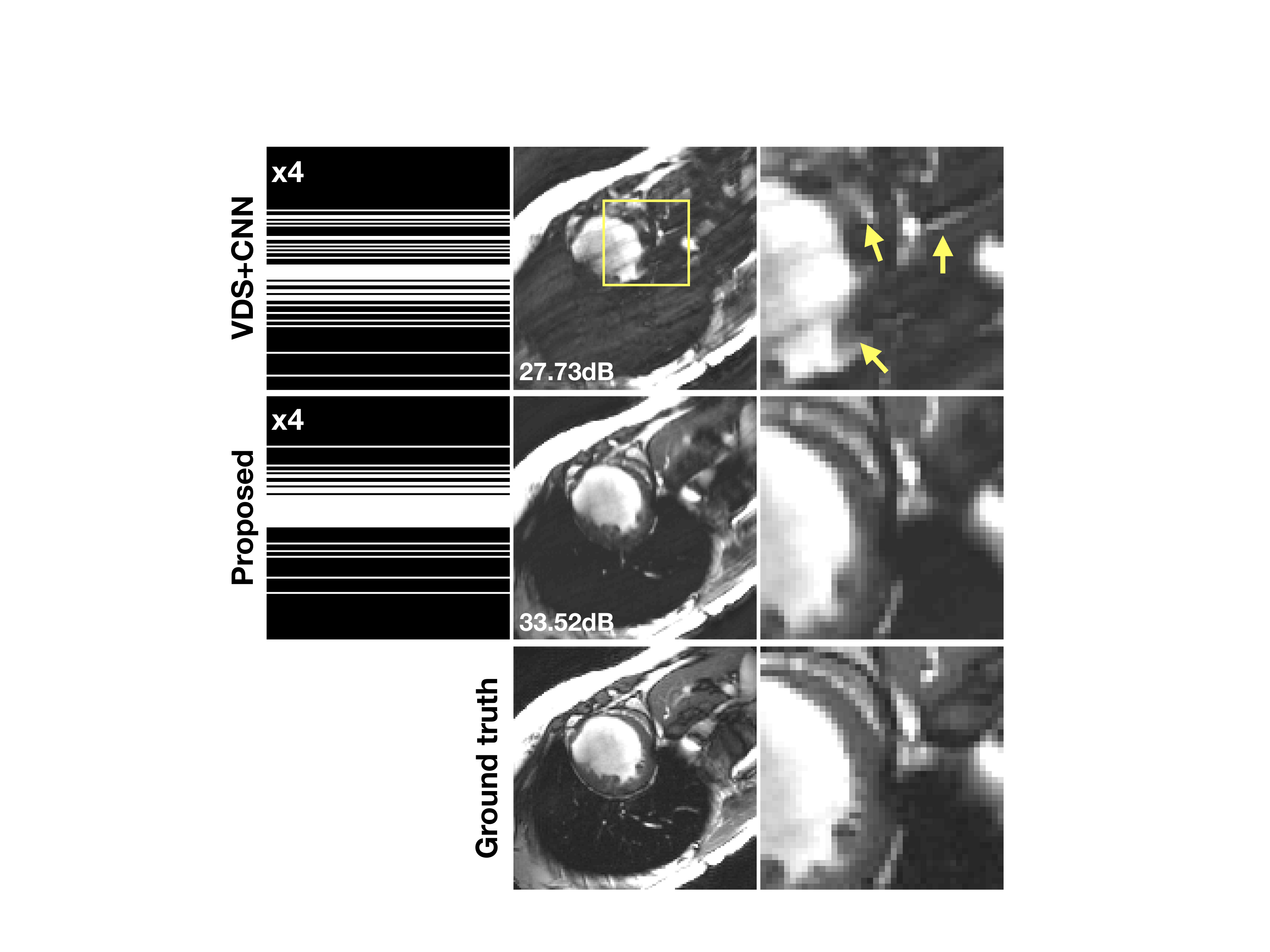}
\vspace{-7mm}
\caption{Reconstruction using (top) the method of 
 variable density sampling (VDS) \cite{he2016deep}
, (middle) our method, compared
    to (bottom) the ground-truth image. (left) sampling patterns,
    (middle) reconstruction result, (right) zoom-in of the region
    denoted with the rectangle in the middle. By learning to jointly 
    sample and reconstruct, our method gives a
    PSNR of 33.52 dB, to be compared to 27.73 dB in \cite{he2016deep}. }
\label{fig:teaser}
\end{figure}


To obtain high-quality MRI within a limited sampling budget, researchers have
investigated compressed-sensing-based methods~\cite{lustig2007sparse,5484183}. This problem,
often referred to as accelerated MRI, is traditionally tackled by exploiting
the fact that MRI images, even when sampled fully, are sparse in nature
  once transformed into, for example,
  wavelets~\cite{lustig2007sparse,guerquin2011fast}, and by considering
  the statistics of the target data~\cite{ravishankar2011mr}. Recently, as in
many other areas, deep-learning-based methods have shown promising
results~\cite{schlemper2018deep,jin2017deep,lee2017deep,8434321}.  However, these
methods mainly focus on the reconstruction part of the problem and
assume a fixed sampling pattern. In other words, they do not consider how
  data should be acquired and are therefore passive in nature.
  
As the acquisition pattern plays a critical role in the final reconstruction
quality, researchers have also focused on optimizing the sampling
pattern in k-space. For example, compressive-sensing-based techniques have
shown promising outcomes with variable-density sampling~\cite{8382318}. The  work in~\cite{gozcu2018learning} goes beyond
variable-density patterns by taking advantage of the statistics of the data.  
However, when learning
or deriving these sampling patterns, existing methods consider the
reconstruction process as fixed and, therefore, disregard the
intertwined nature of the problem.

A possible reason why existing methods do not consider the two problems
together is that it is not straightforward to incorporate the sampling design
in an optimization framework.  Sampling is a categorical variable and is not
differentiable, which prevents the use of simple gradient-based methods.  An
exhaustive search is inaccessible due to the size of as the solution space. 
Greedy methods like \cite{gozcu2018learning} are not viable
either because they would require one to investigate too many cases, considering
that both sampling and reconstruction are optimized jointly.
  
In this paper, we incorporate data acquisition 
(sampling) into our learning framework and to learn to jointly sample and
reconstruct. By doing so, our method learns to
actively obtain data for high-quality reconstruction and is able to go
beyond what can be done by doing either separately.
Specifically, we propose to simultaneously train two neural networks, each
dedicated to reconstruction and to sampling.  To incorporate the sampling process
in our training setup, we take inspiration from
\cite{silver2017mastering}. We learn a sampling network that determines the
next sampling position, in other words, we learn a progressive sampler.  The training
supervision for this network is self-generated through Monte Carlo tree search
(MCTS)~\cite{kocsis2006bandit,coulom2006efficient}, where the search is guided
by the current sampling network and results in a sampling pattern that is
better than what is possible with the sampling network alone.  However, unlike
\cite{silver2017mastering}, where a network is used in place of a rollout, we
perform rollout as in the original MCTS and use the peak signal to noise ratio
(PSNR) given by the reconstruction network linked with the
performance of the reconstruction network.  In case of the reconstruction
network, we train it with the signal sampled through MCTS as input and the
ground-truth fully sampled image as the desired output.  At test time, the
policy network progressively generates the sampling pattern by looking at the
current reconstruction of the network.  As shown in \fig{teaser}, this allows
our network to outperform the state of the art.  To the best of our knowledge,
our work is the first successful attempt at learning the accelerated MRI
pipeline as a whole.

The remainder of the paper is as follows: We first discuss related works in
\secref{related}. We then formalize the problem in \secref{problem}. In
\secref{method}, we explain our method, and in \secref{impl} we provide
implementation details. We report experimental results in \secref{result} and
conclude in \secref{conclusion}.


\section{Related Works}
\label{sec:related}

\begin{figure}
\centering
\includegraphics[trim = 68mm 40mm 70mm 55mm,clip=true,width=\linewidth]{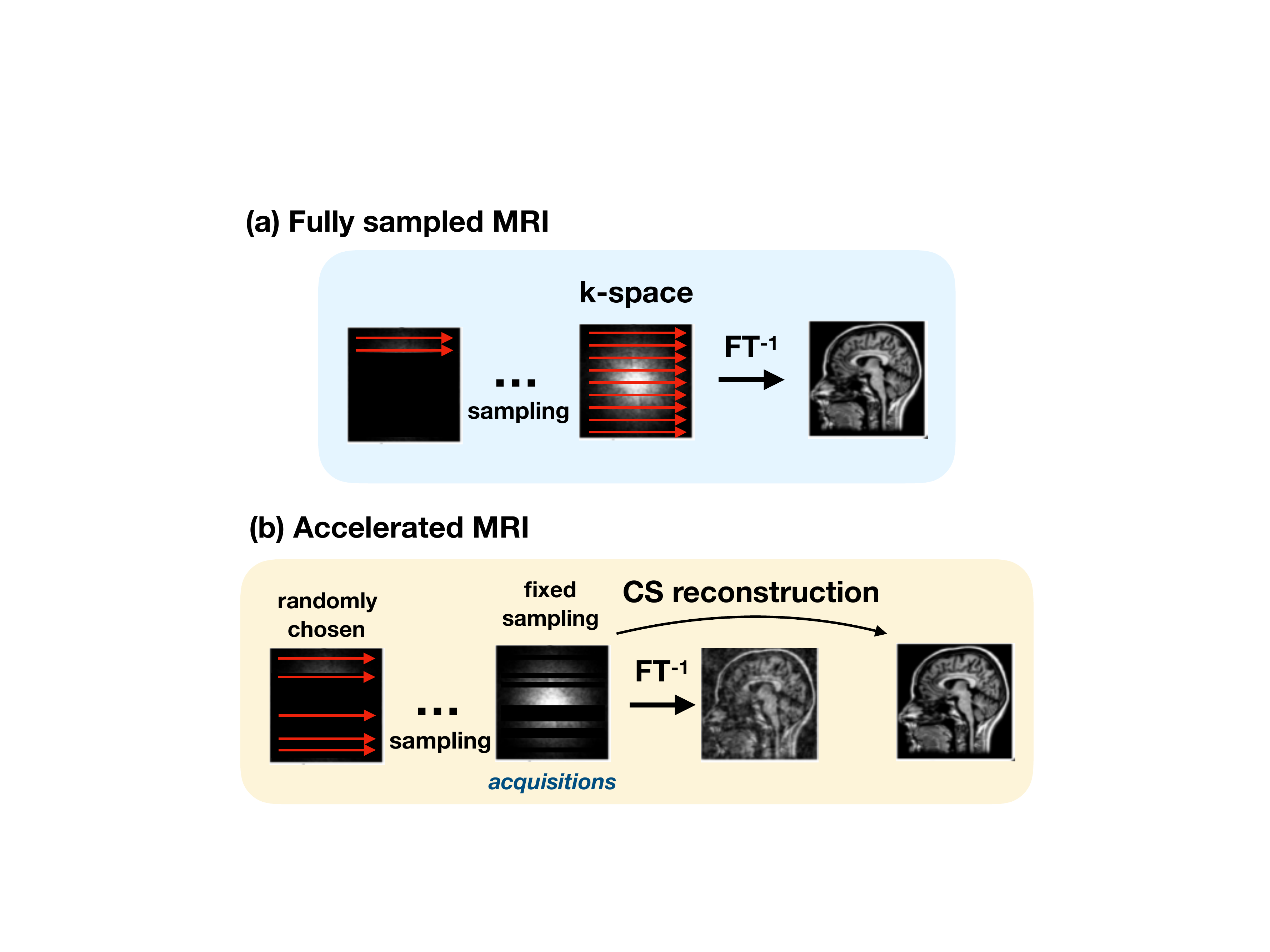}
\vspace{-5mm}
\caption{Accelerated MRI. (a) Fully sampled MRI
  collects all k-space lines to obtain a target image. 
  (b) Accelerated MRI acquires fewer k-space lines and then reconstructs
  downsampled acquisitions from an aliased image. 
  Compressed-sensing (CS) reconstructions rely
  on the sparsity of the image domain.}
\label{fig:csmri}
\end{figure}

As shown in \fig{csmri}, for accelerated MRI, one samples k-space partially, but then reconstructs the images to higher resolution.
Since the pioneering work of~\cite{lustig2007sparse}, a large body of works
exists for accelerated MRI. Here, we first present the compressive
sensing for accelerated MRI, then works related to learning optimal
sampling patterns, and, finally, works that focus on the reconstruction phase of
the pipeline.

\subsection{Compressive Sensing in Accelerated MRI}
In~\cite{lustig2007sparse}, compressed
sensing~\cite{candes2006robust,donoho2006compressed} is applied to reconstruct
sub-Nyquist measurements.  The premise behind the reconstruction process is
that typical MRI images have sparse coefficients in some transform domain, such
as wavelets \cite{lustig2007sparse}, and can therefore be iteratively
reconstructed with a sparsity constraint as prior~\cite{candes2006robust}.  By
doing so, the authors achieve a faster acquisition time and an alias-free image
reconstruction under random sampling. Specifically, in~\cite{lustig2007sparse},
random variable-density sampling is used to sample the low-frequency components
more densely and improve the image quality.

Since its first introduction, several variable-density sampling methods have
been proposed. In~\cite{vasanawala2011practical}, a Poisson-disc distribution
is exploited to reduce correlated effects caused by closely located samples
while retaining a uniform distance between samples.
In~\cite{chauffert2014variable}, continuous trajectories have been proposed.
In~\cite{adcock2015quest}, the authors argue that taking into account both
shared and non-shared components when sparsifying is important when
reconstructing.

Those sampling and reconstruction methods have all shown superior quality over
\cite{lustig2007sparse}.  However, they all suffer from the fact that the
optimal choice of the variable density depends highly on the target dataset.
Therefore, manual intervention per dataset and, sometimes, per image, is
necessary if one is to obtain good reconstruction quality, making it less
practical.

\subsection{Learned Sampling patterns for Accelerated MRI}

Beyond handcrafted sampling patterns, learning-based methods have shown
promising results.  One of the very first method to do so successfully is
\cite{ravishankar2011adaptive}, where the authors define finite non-overlapped
cells with a certain width in k-space and then switch their locations according
to reconstruction performances in a greedy fashion based on a sorted list from
infinite-\textit{p} norm.  To evaluate the reconstruction, they rely on a
dictionary-based reconstruction ~\cite{ravishankar2011mr}. Their method
performs better than previous compressed-sensing methods but suffers from two
drawbacks. One is that the optimal choice of \textit{p} for the
infinite-\textit{p} norm depends highly on the dataset. The other is that there
is no guarantee that the pretrained dictionary is applicable to the data in
question.

Instead of an infinite-\textit{p} norm and pretrained dictionaries, researchers
have also considered the benefit of working directly with the k-space power
spectrum~\cite{knoll2011adapted,vellagoundar2015robust}. A standard assumption
is that the k-space components that hold more power are more important than the
ones with less power.  It leads to a selection of the sampling pattern based on
the k-space energy distribution from the reference. Again, these methods suffer
from the fact that there is no guarantee that the power spectrum of a reference
would be similar to testset, as the selection is performed
off-line. Furthermore, as we show later, a simple sampling of the high-energy
signals does not necessarily yield better reconstructions.

As another direction of research, the relationship between the forward model
matrix and trusted sampling positions \cite{8382318,krahmer2014stable} has been
investigated. A performance guarantee based on basis components of the forward
model has been derived.  Still, these approaches have a limited consideration
on signal energy distribution of each dataset.

Recently, methods based on statistical theory have gained interest.  The
authors of \cite{gozcu2018learning} use the PSNR or the structural similarity
index (SSIM)~\cite{wang2004image}, which are typical image-quality measures,
and greedily search for the sampling pattern that maximizes either one. Their
method is greedy and progressive. It looks at how the measure of interest
evolves when a new sample position is added to the sampling pool. Statistical
error bounds for the optimal estimation of sampling patterns are provided in
the form of theorems.  They require the reconstruction pipeline to be fixed
throughout the greedy selection process; but there is no guarantee that the
greedy selection would lead to an optimal solution.

\subsection{Neural Networks for Accelerated MRI}

Recently, as in many other areas, deep learning has become a popular approach
for accelerated MRI reconstruction.  Typically, convolutional neural networks
(CNN) are trained in a fully supervised manner to regress to high-resolution
reconstructions~\cite{schlemper2018deep,jin2017deep,lee2018deep,doi:10.1002/mrm.26977}.
In \cite{jin2017deep}, multi-resolution CNN with residual learning enhances the
quality of accelerated MRI by encapsulating physical models before input is
fed.  In \cite{schlemper2018deep} and \cite{doi:10.1002/mrm.26977}, cascaded
chains of CNN with data consistency constraints yield performance gain over
dictionary-based reconstructions~\cite{ravishankar2011mr}.  In
\cite{lee2018deep}, inherently complex MRI are split into magnitude and phase
components. For each component, a distinct network is trained to remove
distortions from acceleration.  The author of \cite{lee2018deep} have shown
promising results and achieved higher reconstruction quality over traditional
methods. Beyond simple regression, generative adversarial networks have also
been recently exploited to avoid over-fitting~\cite{yang2018dagan} .

While effective, existing methods still rely on fixed sampling patterns for
data acquisition. As we illustrate in \secref{result} through experiments, this
limits the capabilities of these advanced reconstruction methods.


\section{Problem Formulation}
\label{sec:problem}

Before we detail our method, we first formalize the accelerated MRI
problem. Throughout the formalization, for simplicity, we assume a finite,
discrete, and complete signal model. We write the vectorized full-resolution
signal in the image domain as $\bx \in \mathds{C}^{N}$, where $N$ is the number
of pixels in the image.

We denote the discrete Fourier transform by the matrix
$\bF\in \mathds{C}^{N\times N}$.  For accelerated MRI, not all frequencies are
sampled.  Denoting the sampling process by the Boolean matrix
$\bP \in \mathds{R}^{M\times N}$, where $M < N$ is the number of samples, the
raw measurements $\by$ satisfy
\begin{equation}
  \by = \bP\bF\bx.
  \label{eq:sampling}
\end{equation}

Given \eq{sampling}, we want to obtain an accurate estimate of $\bx$ by using
the observation $\by$.  Formally, if we denote the reconstruction process as
$f_{\btheta}\left(\cdot\right)$, where $\btheta$ is the vector of parameters
that define the reconstruction process, and a quality measure, for example the
PSNR, Euclidean distance, or SSIM~\cite{wang2004image}, as
$g\left(\cdot, \bx\right)$, then the problem of learning a system for
accelerated MRI can be formulated as the problem of finding $\bhtheta$ and
$\bhP$ such that
\begin{equation}
  \label{eq:prob}
  \bhtheta, \bhP = \argmax_{\btheta, \bP} \mathbb{E}_\bx
  \left[
    g \left(
      f_{\btheta}\left(\bP\bF\bx\right), \bx
    \right)
  \right],
\end{equation}
where $\mathbb{E}_\bx$ denotes expectation over $\bx$.  This formulation
involves both $\btheta$ and $\bP$. It is therefore a problem that involves both
sampling and reconstruction. What makes it difficult is that $\bP$ is Boolean,
which leads to a combinatorial optimization.

Existing works can also be understood in the formulation \eq{prob}. In the
traditional compressive-sensing setup~\cite{lustig2007sparse}, $\bP$ would be a
fixed realization of a random-variable density.  For the reconstruction part
driven by $f_{\btheta}\left(\cdot\right)$, it would be a nonparametric process
that involves the solution of a sparse optimization problem. Formally,
\begin{equation}
  \begin{aligned}
    f\left(\by\right) = & \argminS_{\bx} \| \bPsi \bx \|_{\ell_1} \\
    & \text{subject to}\;\; \by=\bP \bF \bx,
  \end{aligned}
  \label{eq:cs_recon}
\end{equation}
where $\bPsi$ is a sparsifying transform, for example the wavelet transform or
the finite differences corresponding to total variation (TV).  In the case of
deep-learning-based methods~\cite{schlemper2018deep,jin2017deep,lee2018deep},
$\bP$ would still remain fixed and $f_{\btheta}\left(\cdot\right)$ would be a
deep network with $\btheta$ now being the parameters of the network. Thus, one
would only optimize for $\btheta$ in \eq{prob}.  Conversely, in works related
to finding the optimal sampling
pattern~\cite{gozcu2018learning,ravishankar2011adaptive} represented by the
optimal sampling matrix $\bP$, the reconstruction process remains fixed and,
therefore, one optimizes only for $\btheta$.

Unlike the existing methods, we optimize jointly $\btheta$ and $\bP$.  This
allows our method to outperform what can be done by training them separately.


\section{Method}
\label{sec:method}

\begin{figure}[t]
\centering
\includegraphics[trim = 80mm 40mm 80mm 60mm,clip=true,width=0.98\linewidth]{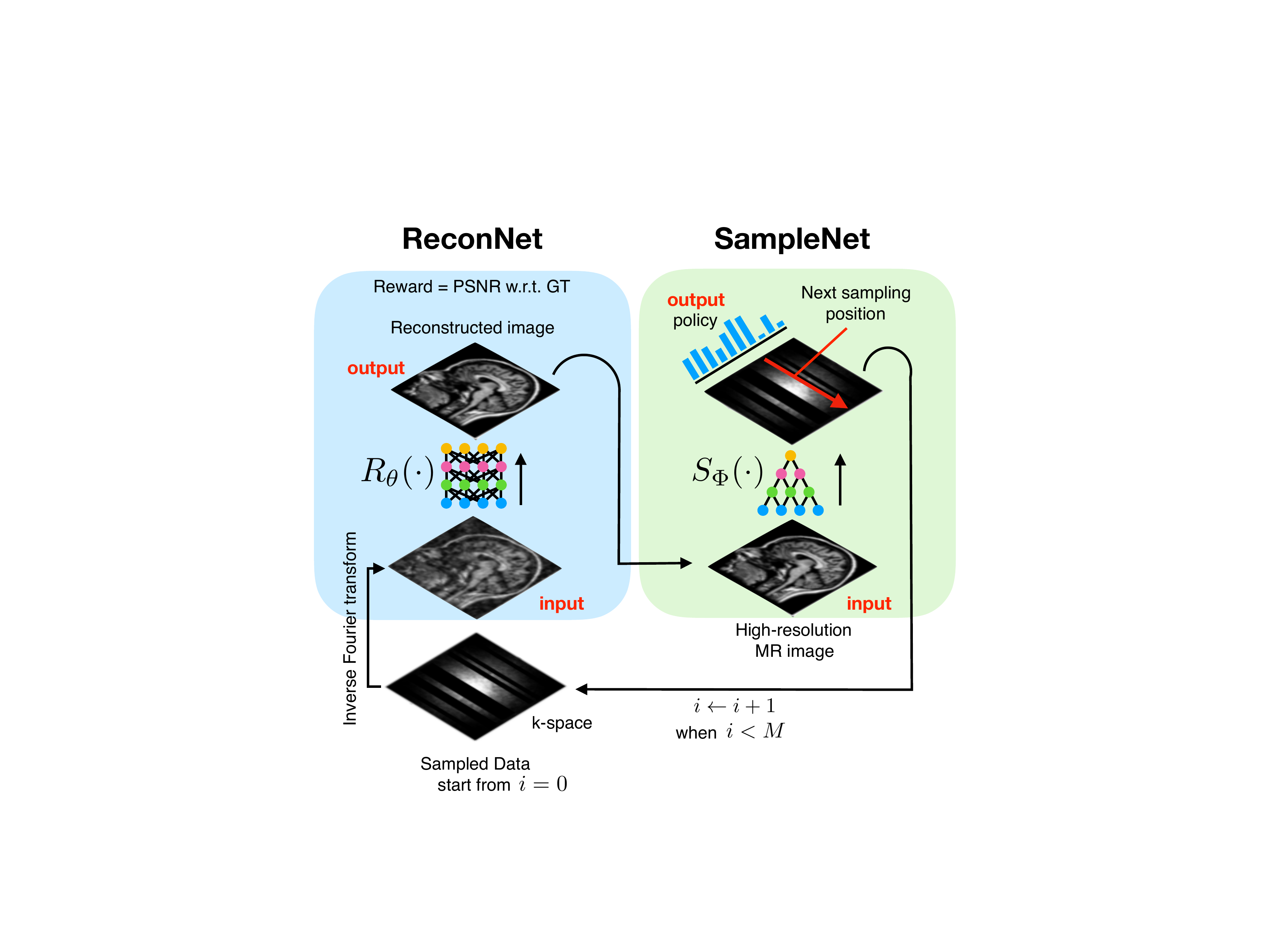}
\vspace{-5mm}
\caption{Overall framework of our method. We train two deep neural
  networks. One learns to reconstruct a high-resolution MRI, and the other
  learns to estimate the policy for determining the position of the next sample.
  We progressively
    sample, with SampleNet, based on the
    reconstructed outcome of ReconNet using collected data.
  }
\label{fig:concept}
\end{figure}


We first provide an overview of the architecture and the training framework. We
further detail how self-supervision signals are obtained through MCTS.

\subsection{Overall Framework}
\label{sec:framework}

In order to learn to jointly sample and reconstruct, as shown in \fig{concept},
we learn two deep networks that are tied together. One network learns to
reconstruct a high-resolution image given the sampled data.  The other learns
to estimate the position of the next sample, given the previous reconstruction
result. We refer to the former as ReconNet and to the latter as SampleNet.

\subsubsection{ReconNet}
In more details, in case of ReconNet, as in other recent deep-learning-based
methods for reconstruction~\cite{jin2017deep,lee2017deep}, we operate in the
image domain instead of the k-space domain, starting from the backprojection
$\bF^{-1}\bP^{\top}\by$.If we denote the reconstruction network as
$R_\btheta\left(\cdot\right)$, then the reconstruction achieved by ReconNet is
\begin{eqnarray}
  f_{\btheta}\left(\by\right)
  &=& R_{\btheta}\left(\bF^{-1}\bP^{\top}\by\right) \nonumber\\
  &=& R_{\btheta}\left(\bF^{-1}\bP^{\top}\bP\bF\bx\right). 
      \label{eq:recon}
\end{eqnarray}
A $\bP$ is a matrix of ones and zeros that represents sampling, the
multiplication with $\bP^{\top}$ undoes the sampling process by padding the
unobserved signal with zeros.

\subsubsection{SampleNet}
For SampleNet, we opt for a progressive-sampling strategy where we draw one
sample at a time using all data acquired up to the instant we sample.  In other
words, the learning is such that SampleNet outputs a probabilistic policy for
determining the next sampling point given the current reconstruction. SampleNet
would be called a policy network in reinforcement-learning
literature~\cite{silver2017mastering,silver2016mastering}. At sampling time
$t$, if we denote SampleNet $S_\bPhi\left(\cdot\right)$ with parameter $\bPhi$
and the probabilistic policy as $\bpi_{t}$, then, using \eq{recon}, we write
that
\begin{equation}
  \bpi_t = S_{\bPhi}\left(
    R_{\btheta}\left(
      \bF^{-1}\bP_t^{\top}\bP_t\bF\bx
    \right)
  \right).
  \label{eq:samplenet}
\end{equation}
We can then write a recursive formula for the sampling pattern at time $t$ as
\begin{equation}
  \bP_{t+1} = 
  \left[
    \begin{array}{c}
      \bP_t\\
      \sel_{\argmax{\bpi_t}}
    \end{array}
  \right],
  \label{eq:pattern}
\end{equation}
where $ \left[
  \begin{array}{c}
    \bA\\
   \bB 
  \end{array}\right]$ denotes concatenation in column direction between $\bA$ and $\bB$, and
$\sel_{\argmax{\bpi_t}}\in\mathds{R}^{N}$ is a Boolean vector with all values
$0$ except for a single element that is $1$ at $\argmax{\bpi_t}$.

With Eqs.~\eqref{eq:recon},~\eqref{eq:samplenet}, and~\eqref{eq:pattern}, we
can now rewrite the learning problem in \eq{prob} as an optimization problem
that involves the training of two networks. We write that
\begin{equation}
  \bhtheta, \bhPhi = \argmax_{\btheta, \bPhi} \mathbb{E}_{\bx, t}
  \left[
    g \left(
      R_{\btheta}\left(\bF^{-1}\bP_t^\top\bP_t\bF\bx\right), \bx
    \right)
  \right].
  \label{eq:prob_our}
\end{equation}
The expectation is now also on $t$, as we are performing progressive sampling,
and the two networks should also be able to deal with various sampling times.

This formulation, however, cannot be solved with conventional gradient-based
methods because \eq{prob_our} is non-differentiable. Instead, we propose to
train the two networks in an alternative way through self-supervision.


\subsection{Self-Supervised Learning}
\label{sec:training}

\begin{figure*}
\centering
\includegraphics[trim = 40mm 50mm 40mm 50mm,clip=true,width=0.9\linewidth]{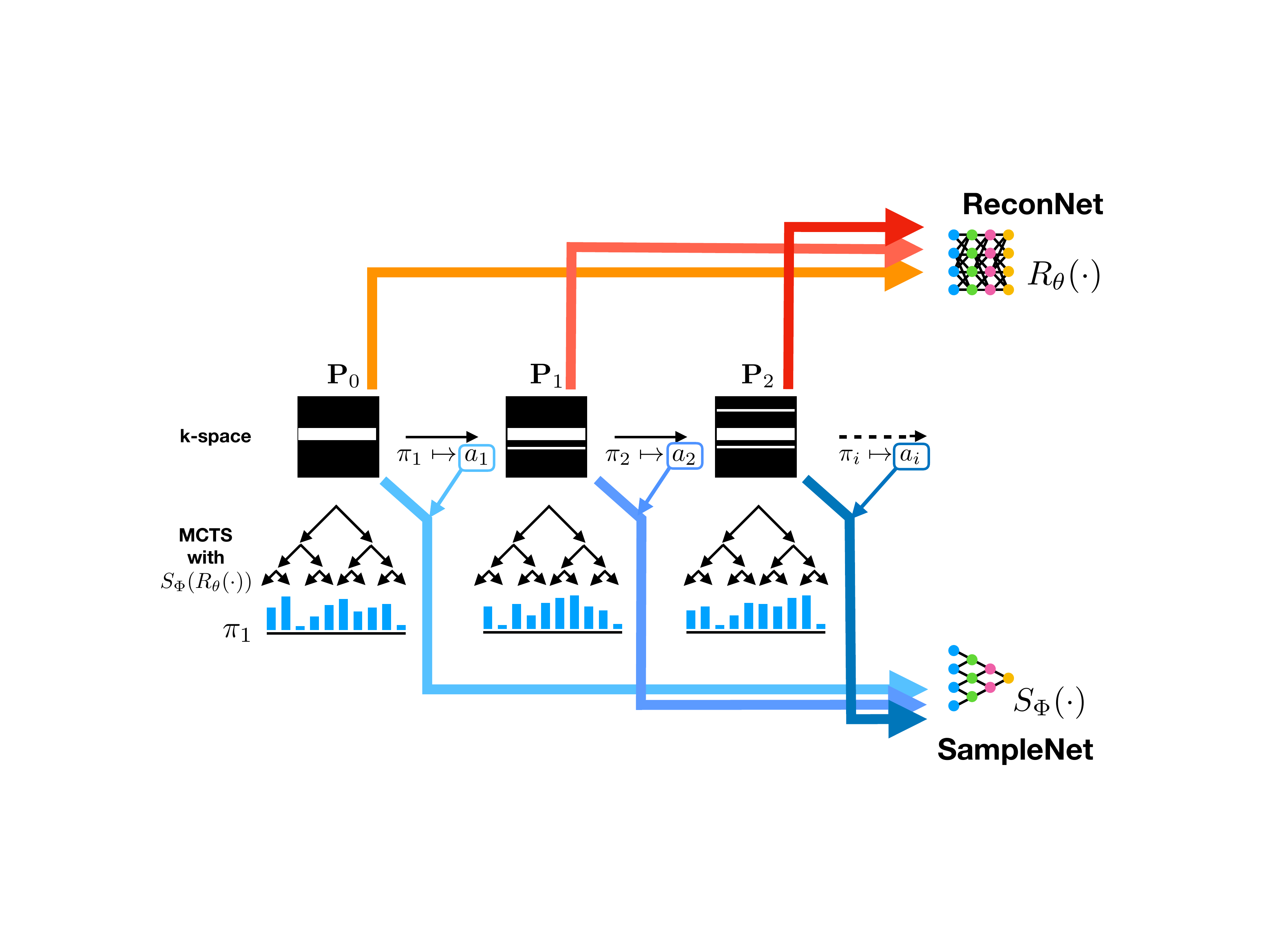} 
\vspace{-4mm}
\caption{  Our   training   self-supervised   learning
  framework. At each sampling round, MCTS returns
  a policy distribution using both the reconstruction network (ReconNet) and
  the sampling network (SampleNet). A sample position is chosen from this
  distribution and sampled. We then proceed to the next sampling
  round. A single search tree is kept throughout the entire sampling
  process. Afterwards, we train the ReconNet to reconstruct high-quality signals
  using these sampling patterns and data, and  SampleNet to emulate the
  final outcome of MCTS.
 }
\label{fig:training}
\end{figure*}


To train ReconNet and SampleNet, we draw inspiration from
AlphaGo~\cite{silver2017mastering,silver2016mastering}.  We propose to train
our two networks with their own separate objectives, which will eventually lead
to satisfying the original objective. Conceptually, we train ReconNet so that
it learns to improve on a given sampling pattern, and also learn SampleNet so
that it learns to provide better sampling patterns given the reconstruction,
which will then result in consistently better final reconstruction quality.

In more detail, as shown in \fig{training}, we rely on
MCTS~\cite{kocsis2006bandit,coulom2006efficient} for the improvement of these
networks through the optimization rounds. We explain in detail how we use MCTS
in \secref{mcts}. For now, assume that MCTS is a black-box component that gives
a sample policy $\tilde{\bpi}_{t}$ that improves on $\bpi_{t}$, the policy from
SampleNet with current network parameters $\btheta_n$ and $\bPhi_{n}$, where
$n$ is the round of optimization. Formally, we write that
\begin{equation}
  \label{eq:pi_mcts}
  \tilde{\bpi}_{t} = \MCTS\left(\bP_{t}, \bx, \btheta_n, \bPhi_n\right).
\end{equation}
Then, with MCTS, we create sampling patterns $\tilde{\bP}_{t}$ recursively by
performing
\begin{equation}
  \tilde{\bP}_{t+1} = 
  \left[
    \begin{array}{c}
      \tilde{\bP}_{t}\\
      \sel_{\argmax{\tilde{\bpi}_{t}}}
    \end{array}
  \right],
  \label{eq:P_mcts}
\end{equation}
which leads to better reconstruction results
\begin{equation}
  \label{eq:x_mcts}
  \tilde{\bx}_{t} = 
  R_{\btheta_{n}}\left(
    \bF^{-1}\tilde{\bP}_{t}^{\top}\tilde{\bP}_{t}\bF\bx
  \right).
\end{equation}
With $\tilde{\bP}_{t}$, $\tilde{\bpi}_{t}$, $\tilde{\bx}_{t}$, and $\bx_{t}$,
we train both ReconNet and SampleNet.

For ReconNet, we train to obtain high-quality reconstructions, given the
sampling patterns from MCTS. In other words, we simply replace $\bP_{t}$ with
$\tilde{\bP}_{t}$ in \eq{prob_our}. By doing so, we remove the dependency of
$\bPhi$, leading to the per-round optimization
\begin{equation}
  \label{eq:opt_recon}
  \btheta_{n+1} = 
  \argmax_{\btheta | \btheta_0=\btheta_{n}} \mathbb{E}_{\bx, t}
  \left[
    g \left(
      R_{\btheta}\left(\bF^{-1}\tilde{\bP}_{t}^\top\tilde{\bP}_{t}\bF\bx\right), \bx
    \right)
  \right],
\end{equation}
where $\btheta_0$ is the starting point of the optimization. This can now be
trained with a gradient-based solver such as ADAM~\cite{Kingma15}.  The
standard PSNR measure in Eq. \eqref{eq:psnr} is used for $g$.

We train SampleNet to emulate the outcome $\tilde{\bpi}_{t}$ of MCTS.  Thus,
with \eq{samplenet}, and denoting cross entropy as $H\left(\cdot,\cdot\right)$,
we write that
\begin{equation}
  \label{eq:opt_sample}
  \bPhi_{n+1} = \argmin_{\bPhi | \bPhi_0 = \bPhi_n} \mathbb{E}_{\bx, t}
  \left[
    H \left(
      S_{\bPhi}\left(
        \tilde{\bx}_{t}
      \right),
      \tilde{\bpi}_{t}
    \right)
  \right],
\end{equation}
where $\bPhi_0$ is the initial value of $\bPhi$ for the optimization. This
objective can also be optimized through gradient descent.

In practice, to avoid overfitting, we optimize \eq{opt_recon} and
\eq{opt_sample} for only $K$ steps per round.  We also apply experience replay,
by creating the training batch from randomly selected past examples. This
further prevents overfitting and ensures smooth
optimization~\cite{silver2017mastering,silver2016mastering}.  A summary of the
process is provided in Algorithm~\ref{alg:training}, with subscripts for the
optimization round and the reconstruction episode added for clarity.

\begin{algorithm}[t]
\caption{\label{alg:training} Self-supervised learning with Monte Carlo tree
  search and experience replay.}
\begin{center}
\begin{algorithmic}[1]
  \Require {$\{$$\bx$: ground-truth data, $\bPsi$: experience memory,
    $\bP_{0}$: initial sampling pattern, $L$: number of optimization rounds,
    $E$: number of reconstruction episode per optimization round, $T$:
    sampling budget, and $K$: number of optimization steps per round $\}$}
\Function{SelfTrain}{$\bx$, $\bPsi$, $\bP_{0}$}
\For{$n=1$ \TO $L$}
\For{$m=1$ \TO $E$}
\Statex{\hspace{4.5em}// Initialize}
\State{$\tbP_{1} = \bP_0$}
\State{$\MCTS \leftarrow $ Initialize search tree}
\Statex{\hspace{4.5em}// Build self-supervision through MCTS}
\For{$t=1$ \TO $T$}
\State{
  $\tbpi_{t}
  =
  \MCTS\left(\tbP_{t}, \bx_{m,n}, \btheta_{n}, \bPhi_{n}\right)$
  (from \eq{pi_mcts})
}
\State{
  $\tbP_{t+1}
  =
  \left[
  \begin{array}{c}
  \tbP_{t}\\
  \mathds{1}_{\argmax{\tbpi_{t}}}
  \end{array}
  \right]
  $ 
  (from \eq{P_mcts})
}
\State{
  $\tilde{\bx}_{m,t,n}
  =
  R_{\btheta_{e}}\left(
    \bF^{-1}\tbP_{t+1}^{\top}\tbP_{t+1}\bF\bx_{m,n}
  \right)$
}
\State{\hspace{3cm}(from \eq{x_mcts})}
\State{
  $\bPsi \xleftarrow{\textit{append}} \left(
    \tbpi_{t}, \tbP_{t}, \tbx_{m,t,n}, \bx_{m,n}
  \right)$ }
\EndFor
\EndFor
\Statex{\hspace{3.0em}// Train}
\State{$\bPsi_{n} \leftarrow$ Random examples from $\bPsi$}
\State{$\btheta_{n+1} \leftarrow$ Optimize K times with $\bPsi_{n}$ (from \eq{opt_recon})}
\State{$\bPhi_{n+1} \leftarrow$ Optimize K times with $\bPsi_{n}$ (from \eq{opt_sample})}

\EndFor

\State{\textbf{return} $\btheta_{L}$, $\bPhi_{L}$}

\EndFunction
\end{algorithmic}
\end{center}
\end{algorithm}



\subsection{Monte Carlo Tree Search for Sampling}
\label{sec:mcts}

\begin{figure*}
\centering
\includegraphics[trim = 10mm 50mm 10mm
65mm,clip=true,width=0.8\linewidth]{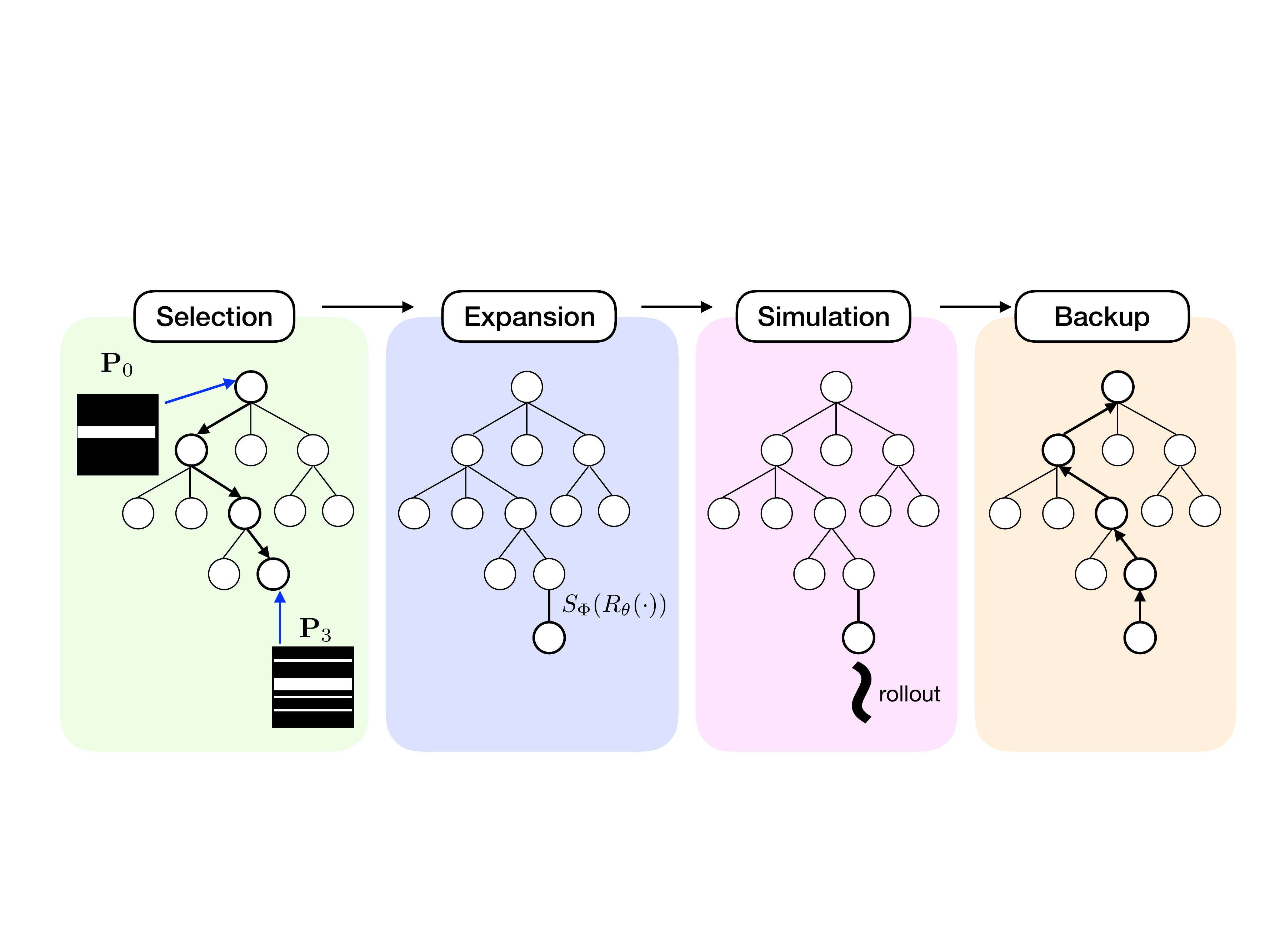} 
\vspace{-4mm}
\caption{Our Monte Carlo tree search. Each node in the tree denotes a sample pattern, while 
    a move down the tree involves sampling a new position. We traverse the tree
    until a leaf node has been reached, where we simulate sampling with the
    sampling network (SampleNet) until some sampling budget has been
    reached. We then use the actual reconstruction performance as the reward,
    which is then backed up to all parent nodes. When performing the backup, we
    save the average and the maximum to consider the best reconstruction
    within all child nodes as well as the average reconstruction
    quality.
}
\label{fig:mcts}
\end{figure*}


The key insight that allows the learning strategy of \secref{training} to work
is that, during training, the sampling results with MCTS should always be
better than what the two networks could provide without, thus providing
guidance. This is possible because MCTS simulates plays according to a search
policy (in our case, sampling using SampleNet and ReconNet) and finds out how
good a random move is by looking at the ultimate outcomes of the
simulations---in our case, the reconstruction results. With this look-ahead
behavior, MCTS arrives at a decision that is better than that of the original
search policy.

As shown in \fig{mcts}, we perform MCTS in the traditional four stages:
selection, expansion, simulation, and backup. The selection follows the
upper-confidence-bound strategy of typical MCTS setups, and the expansion is
guided by SampleNet and ReconNet.  We perform the simulation until we reach the
sampling budget $M$, so that the final reconstruction quality becomes the
reward that is backed up to update the node statistic in the search tree.

In more details, as we are focusing on retrieving the best reconstruction, we
use implicit minimax backups~\cite{Lanctot14} with SampleNet and ReconNet
together guiding the selection and exploration, as in
\cite{silver2017mastering,silver2016mastering}.  Specifically, for each node in
the search tree identified with the sampling pattern $\bP$, we store
$Q\left(\bP\right)$, the average of the rewards of all child nodes,
$V\left(\bP\right)$, the maximum reward of all child nodes, and
$N\left(\bP\right)$, the number of visits to the node. Then, if we denote the
policy for a certain movement $a$ as $\bpi_a$, where $\bpi$ is from
\eq{samplenet}, then we define the upper confidence bound for movement $a$ at
state $\bP$ as
\begin{equation}\label{eq:ucb}
  \begin{aligned}
    U(\bP,a) = & (1-\alpha)Q(\left[
  \begin{array}{c}
  \bP\\
  \sel_{a}
  \end{array}
  \right]) + \alpha V(\left[
  \begin{array}{c}
  \bP\\
  \sel_{a}
  \end{array}
  \right]) \\
    & + C_{puct}(\left(1-\epsilon\right)\bpi_{a} +
    \epsilon\bdelta)\sqrt{\frac{N(\bP)}{N(\left[
  \begin{array}{c}
  \bP\\
  \sel_{a}
  \end{array}
  \right])}}
  \end{aligned}
  \;\;,
\end{equation}
where $\alpha$ and $C_{puct}$ are hyper-parameters that control whether the
search favors the average or the maximum and control the degree of
exploration performed by MCTS, and
$\bdelta \sim Dir\left(0.3\right)$ is the Dirichlet noise to encourage
exploration as in~\cite{silver2017mastering}, while $\epsilon$ is the
hyper-parameter that controls the degree of exploration. Note that the Dirichlet
noise is applied whenever $\bpi$ is used.  Algorithm~\ref{alg:mcts} is a
summary of the entire MCTS process. We detail the hyper-parameter settings in
\secref{train_setup}.

In \cite{silver2017mastering}, a deep network that estimates the
reward---referred to as the value network---was used instead of simulation due to 
the long nature of Go games. However, we found that the use of a value network gives
worse reconstructions as shown in the experiments of \secref{result}.

\begin{algorithm}[t]
  \caption{ Monte Carlo tree search with implicit minimax backup. }
  \label{alg:mcts}
  \begin{algorithmic}[1]
    \Require{$\{$
      $\bP$: sampling pattern,
      $\bx$: ground-truth data,
      $\btheta$: parameters of the reconstruction network,
      $\bPsi$: parameters of the sampling network,
      $Q\left(\cdot\right)$: average reward of all child nodes,
      $V\left(\cdot\right)$: maximum reward of all child nodes,
      $N\left(\cdot\right)$: number of visits to a node,
      $T$: sampling budget, and
      $M$: number of MCTS rounds$\}$
    }
    \Function{Search}{$\bP$, $\bx$, $\btheta$, $\bPsi$, $Q\left(\cdot\right)$, $V\left(\cdot\right)$, $N\left(\cdot\right)$}
    \Statex{\hspace{1.5em}// Depending on the availability of sampling budget}
    \If{$\left|\bP\right| < T$}
      \Statex{\hspace{3.0em}// Select next sample}
      \State{$\widehat{\bP}  =
      \left[
  \begin{array}{c}
  \bP\\
  \sel_{a}
  \end{array}
  \right]$}
      \If{$N\left(\bP\right) > 0$}
      \Statex{\hspace{4.5em}// New node, simulate}
      \State{$\bpi \leftarrow S_\bPhi\left(R_{\btheta}\left(\bF^{-1}\widehat{\bP}^{\top}\widehat{\bP}\bF\bx\right)\right)$}
      \State{$\bdelta \sim Dir\left(0.3\right)$}
      \State{$a \sim \left(1 - \epsilon\right) \bpi + \epsilon\bdelta$}
      \Else
      \Statex{\hspace{4.5em}// Existing node, traverse down tree}
      \State{$a \leftarrow \argmax_{a}U\left(\bP, a\right)$ (from \eq{ucb})}
    \EndIf
    \Statex{\hspace{3.0em}// Recursively evaluate next sample}
    \State{\Call{Search}{$\widehat{\bP}$, $\bx$, $\btheta$, $\bPsi$, $Q\left(\cdot\right)$, $V\left(\cdot\right)$, $N\left(\cdot\right)$}}
    \Else
      \State{$\widehat{\bP}  = 
      \left[
  \begin{array}{c}
  \bP\\
  \sel_{a}
  \end{array}
  \right]$}
    \Statex{\hspace{3.0em}// Use actual reconstruction result}
    \State{$\bx_{i} \leftarrow R_{\btheta}\left(\bF^{-1}\widehat{\bP}^{\top}\widehat{\bP}\bF\bx\right)$}
    \State{$v \leftarrow g(\bx_{i},\bx)$}
    \EndIf
    \Statex{\hspace{1.5em}// Backup }
    \State{$ Q(\bP_{i+1}) \leftarrow  \frac{Q(\bP_{i+1}) \cdot N(\bP_{i+1}) +v}{ N(\bP_{i+1}) +1}$} 
    \State{$ V(\bP_{i+1}) \leftarrow \max( V(\bP_{i+1}),v ) $}
    \State{$ N(\bP_{i+1}) \leftarrow N(\bP_{i+1})+1$}
    \State{\textbf{return} $v$}
    \EndFunction
    \Function{MCTS}{$\bP$, $\bx$, $\btheta$, $\bPsi$}
    \Statex{\hspace{1.5em}// Search tree multiple times}
    \For{$m=1$ \TO $M$}
    \State{\Call{Search}{$\bP$, $\bx$, $\btheta$, $\bPsi$, $Q\left(\cdot\right)$, $V\left(\cdot\right)$, $N\left(\cdot\right)$}}
    \EndFor
      \State{$\widehat{\bP}  = 
      \left[
  \begin{array}{c}
  \bP\\
  \sel_{a}
  \end{array}
  \right]$}
    \Statex{\hspace{1.5em}// Return policy from search outcomes}
    \State{$ \tbpi(a) = \frac{N\left(\widehat{\bP}\right)}{\sum_{a}N\left(\widehat{\bP}\right)}$}
    \State{\textbf{return} $\tbpi$}
    \EndFunction

  \end{algorithmic}
\end{algorithm}






\section{Implementation Details}
\label{sec:impl}

We now discuss the details required to implement the proposed framework into an
actual system for accelerated MRI. An implementation of the method is available
online.\footnote{We shall provide a link to the repository once the paper gets
  accepted.}

\subsection{Sampling with 1D Readout}
\label{sec:1d}

Up until now, for the sake of ease in explanation, we motivated our method with
the case where a single position was sampled in the k-space domain. In
practice, MRI with Cartesian trajectory~\cite{lustig2007sparse} is typically
performed by sampling one line of frequency in the k-space, through a process
called readout.  Therefore, in our implementation, $\bpi$ in \eq{samplenet} is
in fact a vector in $\mathds{R}^{\sqrt{N}}$, and $\sel_{\argmax{\bpi_t}}$ is a
matrix in $\mathds{R}^{\sqrt{N} \times N}$, whose rows each correspond to a
sampling vector that samples elements in the row or column of
$\argmax{\bpi_t}$.

\subsection{Network Architectures}

\begin{figure*}
\centering
\includegraphics[trim = 23mm 90mm 0mm 80mm,clip=true,width=1.05\linewidth]{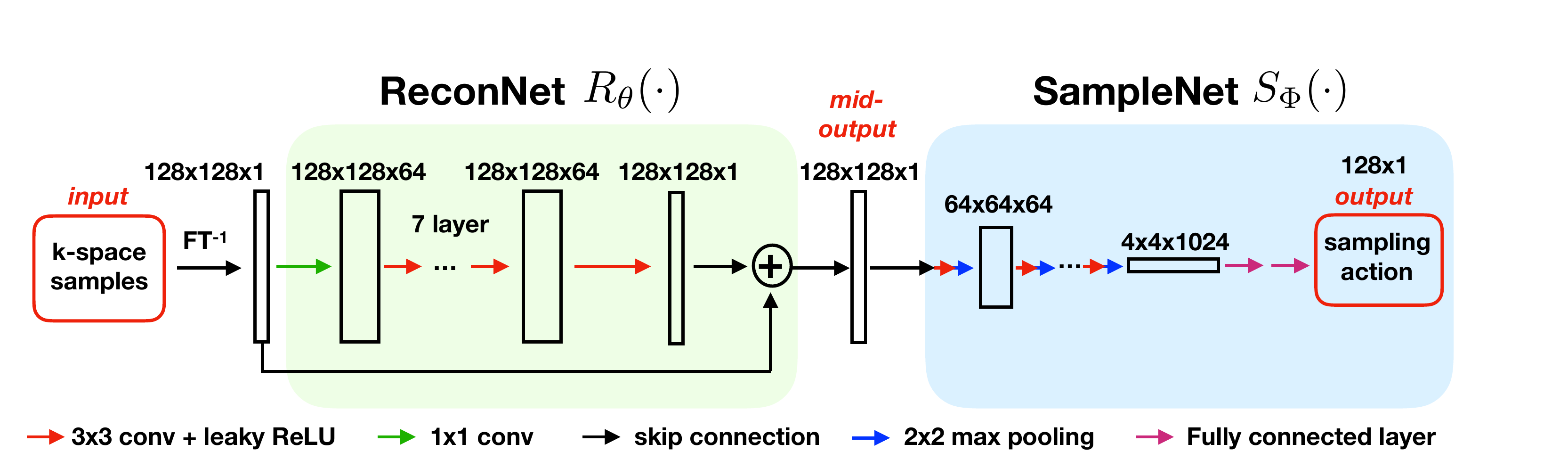} %
\vspace{-8mm}
\caption{Architecture of the two networks. Note that the sampling network
    (SampleNet) takes as input the output of the reconstruction network
    (ReconNet). Both of our networks are based on residual blocks~\cite{he2016deep}.}
\label{fig:architecture}
\end{figure*}


The overall architecture of the two coupled networks is shown in
\fig{architecture}. As discussed in \secref{framework}, ReconNet takes images
in the spatial domain as input, and SampleNet takes the output of ReconNet as
input. Here, the input image is in tensor form to take advantage of existing
deep-learning libraries such as TensorFlow~\cite{Tensorflow}.

\subsubsection{ReconNet}

In order to recover high-quality MRI, we use a fully convolutional network, as
in other recent deep-learning-based
reconstructors~\cite{schlemper2018deep,jin2017deep,lee2017deep,lee2018deep},
but with a simpler architecture. In our early experiments (not shown), we also
employed a more complex U-Net~\cite{Ronneberger15} type of architecture as
in~\cite{jin2017deep}, but achieved similar performance with significantly
higher computation cost. We therefore use the simpler architecture.

We use eight residual blocks~\cite{he2016deep}. Before the residual blocks, we
employ a ($1\times1$) linear convolutional layer to allow the network to easily
adapt to the input data range, if needed. Each block contains a ($3\times 3$)
convolutional layer followed by batch normalization and
leaky-ReLU~\cite{Maas13} activation. All convolutions are zero-padded to have
the same output size as the input, and have 64 output channels. After the
residual blocks, a ($1\times1$) linear convolutional layer is used to convert
the output back to the same number of channels as input.

\subsubsection{SampleNet}

To obtain a probabilistic policy on where to sample next, we use an
architecture that is similar to the policy network in
\cite{silver2017mastering} for our SampleNet. As in ReconNet, each
convolutional block contains a zero-padded ($3\times 3$) convolutional layer
followed by batch normalization and leaky-ReLU activation. After each
convolutional block, we apply ($2\times2$) max-pooling, as well as double the
number of channels until it becomes 256. The first convolutional block has 64
channels. We repeat this structure until the output of the network becomes
($4\times 4 \times 256$). We then flatten and apply a fully connected layer
with $1024$ neurons, again followed by batch normalization and leaky-ReLU
activation. Finally, another fully connected layer with softmax activation is
used to turn the output into a probability. As noted in \secref{1d}, each
element in this probability represents a line.

\subsection{Training Setup}
\label{sec:train_setup}

We use ADAM~\cite{Kingma15} to optimize the networks in
Algorithm~\ref{alg:training}, with the learning rate of $10^{-4}$.  We apply
weight decay constraining the $\ell_2$-norm of the weight parameters of the
network, where the hyper-parameter for the decay is empirically set to
$10^{-4}$.  To prevent the networks from overfitting, we train at max 1000
iterations per round, and at max 3 epochs for the samples in current experience
memory. In addition, we only keep examples from the ten most recent rounds of
optimization.

For MCTS, the number of repetitions of simulation is 10.  The minimax-backup
parameter $\alpha$ is set to $0.5$ and $C_{puct}$ is set to $1.0$.  The
parameter for exploration extent is
$\epsilon=0.25$.

To reduce the time taken to generate supervision, we create these examples in
parallel with 16 threads. We train our model for $540$ rounds. It takes around
four days to train our model on an Intel i7-7820X (3.60GHz) CPU and an NVidia
Titan X (Pascal) GPU, or an Intel Xeon E5-2690 v3 (2.60GHz) CPU and an NVidia
GeForce GTX 1080Ti GPU.


\section{Results}
\label{sec:result}

\subsection{Datasets}

We evaluate our method on two datasets. The first one is composed of cardiac
images, and the second one composed of knee images.

The {\it cardiac} dataset is from the cardiac atlas
project~\cite{radau2009evaluation}.  From this dataset, we discard images that
are corrupted with severe noise. Since the dataset is provided in the spatial
domain, we bring it into the Fourier domain numerically. From the original
image size of $128^2, 256^2$, we retain the central ($128 \times 128$) crop to
avoid the artificial image boundaries that exist due to the characteristics of
the capture devices.  We split the training and test sets based on patients,
where 10 patients are assigned to train, and another 10 patients assigned to
test. This results in a total of 4999 training images and 6963 test images.

The {\it knee} dataset is from an open data
platform\footnote{http://mridata.org/}. The dataset is measured as a 3D volume,
and we use the central 2D slice in our experiments.  The size of the original
images is ($320\times 320$), which we again crop to their central
($128 \times 128$) region to accelerate simulations. The original data is
collected from 8 coils, therefore producing an 8-channel image in k-space,
which we compress down to a single channel to keep the framework identical for
both datasets.  The knee dataset contains fewer images than the {\it cardiac}
dataset. Among the 20 patients, we use 17 patients for training and the
remaining 3 for testing. This amounts to 2550 training images and 450 for
testing. We again keep a small portion of the training data for validation.

For the {\it cardiac} dataset, we use our dual-channel approach to train our
networks while, for the {\it knee} dataset, we use a square-root of
sum-of-squares operation. We empirically found that, for the {\it cardiac}
dataset composed of real-valued images, the dual-channel representation helps:
for the {\it knee} dataset consists of complex-valued images, it does not.

\subsection{Baselines and Evaluation Metric}

\subsubsection{Baselines}

To validate the effectiveness of our method, both SampleNet and ReconNet, as
well as the joint training framework, we compare our method against three
baseline methods:
\begin{itemize}
\item {\bf VDS+TV \cite{becker2011nesta}}:
  Variable density sampling (VDS)~\cite{lustig2007sparse} with reconstruction
  using TV minimization~\cite{becker2011nesta}. This is a standard baseline for
  accelerated MRI.
\item {\bf LCS~\cite{gozcu2018learning}+TV \cite{becker2011nesta}}:
  Learning-based compressive sensing (LCS) from~\cite{gozcu2018learning}, with
  their total-variation reconstruction \cite{becker2011nesta}. We use the
  publicly available source code with tuned
  parameters.
\item {\bf VDS+FBPConv~\cite{jin2017deep}}: A state-of-the-art reconstruction
  pipeline that uses deep CNNs. It uses variable density sampling for the
  sampling pattern.
\end{itemize}
We also include variants using our reconstruction network instead of the
original reconstruction strategies to demonstrate the effectiveness of joint
training.
\begin{itemize}
\item {\bf LPF + Our Recon}: Lowpass filtering (LPF) with our reconstruction
  network. A simple baseline to demonstrate the performance of ReconNet on a
  super-resolution setup.
\item {\bf VDS + Our Recon}: VDS with our reconstruction network.  We use this
  baseline to demonstrate the importance of a sampling pattern when using a
  deep reconstructor.
\item {\bf LCS~\cite{gozcu2018learning} + Our Recon}: We use the sampling
  pattern from \cite{gozcu2018learning} and train ReconNet. We use this
  baseline to show that the pattern learned with a different reconstruction
  method is not the optimal pattern for deep reconstructors.
\end{itemize}
All methods were trained with the same training configurations.  The
downsampling factor equals 4 for all subsequent experiments.

\subsubsection{Evaluation Protocol}

To evaluate the effectiveness of each method, we observe the reconstruction
quality in a scenario where the sample budget is one-fourth of the
full-resolution signal. In other words, we evaluate the case where the
reconstruction needs to upsample by a factor of four.

As evaluation metric, we use the standard PSNR measure 
\begin{equation}\label{eq:psnr}
  \PSNR(\bz,\bx)
  =
  -20\log_{10} \left(
    \frac
    { 1/\sqrt{N} \times \| \bz-\bx\|_2}
    {\|\bx\|_\infty}
  \right),
\end{equation}
where $\by$ and $\bx$ are the reconstructed and reference signals,
respectively.

\subsection{Quantitative Results}

\begin{table*}[!htbp]
  \centering
  \caption{Quantitative results in terms of PSNR for the {\it cardiac} and {\it knee}
    datasets. PSNR of both zero-filled inverse Fourier transform on the sampled
    data and that of using the reconstruction methods are shown. Best results are shown in bold. We show both the results of the
    original method and using our reconstruction network (ReconNet) trained
    with respective sampling patterns. Our method performs best for both
    datasets. On the other hand, our
    method performs well with a single hyper-parameter setup. }
\label{tbl:recon}
\begin{tabular}{@{}
>{\columncolor{white}[0pt][\tabcolsep]}
cccccc
>{\columncolor{white}[\tabcolsep][0pt]}l
@{}}
\toprule
\multicolumn{2}{c}{Sampling}  & VDS & LCS & LPF & Ours \\ \cmidrule{1-2}
Dataset & Recon &  &  &  &  \\
\cmidrule(lr){1-1}\cmidrule(lr){2-2}
\rowcolor{gray!20}
 &zf-IFT & 25.42 & 28.99 & 30.19 & 27.54 \\
\rowcolor{gray!20}&TV & 32.33 & 33.72 & 31.49 & - \\
\rowcolor{gray!20}&FBPConv & 30.21 & 33.77 & 23.79 & -\\
\rowcolor{gray!20} 
\multirow{ -4}{*}{{\it Cardiac}}&Ours & 27.91 & 32.56 & 29.00 & \textbf{34.22}\\
 &zf-IFT & 22.51 & 26.96 & 27.41 & 26.79 \\
&TV & 26.17 & 28.93 & 27.9 & - \\
&FBPConv & 26.55 & 28.95 & 28.04 & -\\ 
\multirow{ -4}{*}{{\it Knee}}&Ours & 25.04 & 28.43 & 28.11 & \textbf{29.1}\\
\bottomrule
  \end{tabular}
\end{table*}


We provide the quantitative results in \tbl{recon}. In addition to the
reconstruction performance, we report the results of simple reconstructions
through an inverse Fourier transform of zero-filled measurements (zf-IFT) to
single out the effect of the sampling pattern.  Our method outperforms all
other compared methods.

As shown by the comparisons with LCS and FBPConv, learning only the sampling
pattern or, simply, the reconstruction network alone, does not provide optimal
performances. Furthermore, by comparing with the variants that use the sampling
patterns discovered by LCS, we see that the simultaneous learning of both
components favorably impacts the performance. Note that our method does not
require per-dataset parameter tuning and is able to outperform the state of the
art with a single hyper-parameter setup. By contrast, methods relying on
TV~\cite{becker2011nesta} require per-dataset tuning.

One interesting thing here is that LPF, which is sampling just the
low-frequency part, gives the highest reconstruction quality. However, training
with this sampling pattern results in over-fitting, as demonstrated by a
decrease in performance when the reconstruction network is added. This is
expected as this sampling pattern does not give any information about the
high-frequency parts of the signal. In a nutshell, the reconstruction network
learns to interpolate the components that were not observed, which there is
nothing to interpolate for this sampling pattern.  However, this is not the
case of the other sampling patterns, as demonstrated by the fact that ReconNet
always improves performance.

Besides outperforming the other baselines, our method is also easily scalable
as it is based on a stochastic gradient-based optimization.

\subsection{Qualitative Results}

\begin{figure*}[t]
\centering
\includegraphics[trim = 0mm 58mm 0mm 50mm,clip=true,width=17 cm]{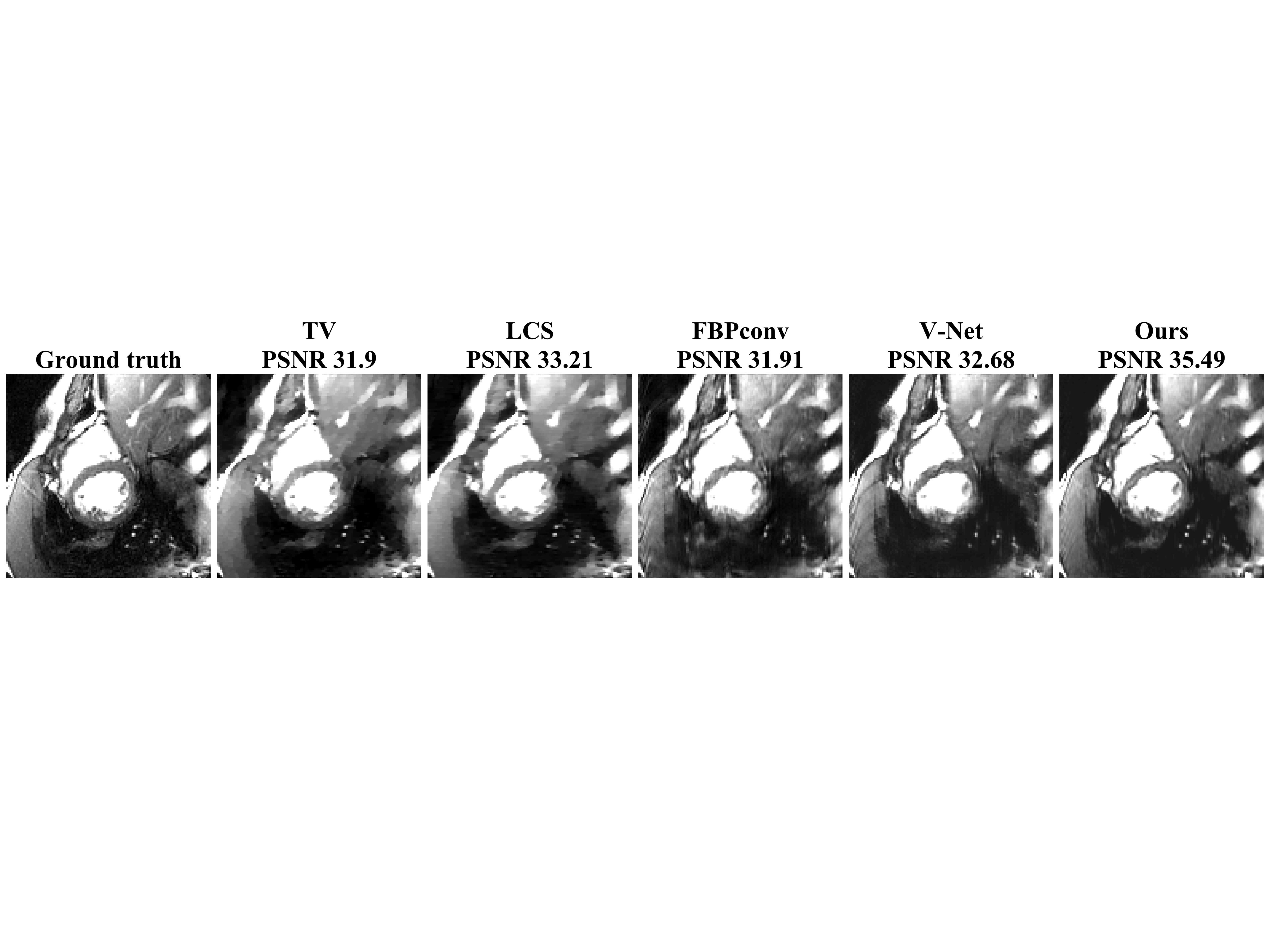}
\includegraphics[trim = 0mm 58mm 0mm 58mm,clip=true,width=17 cm]{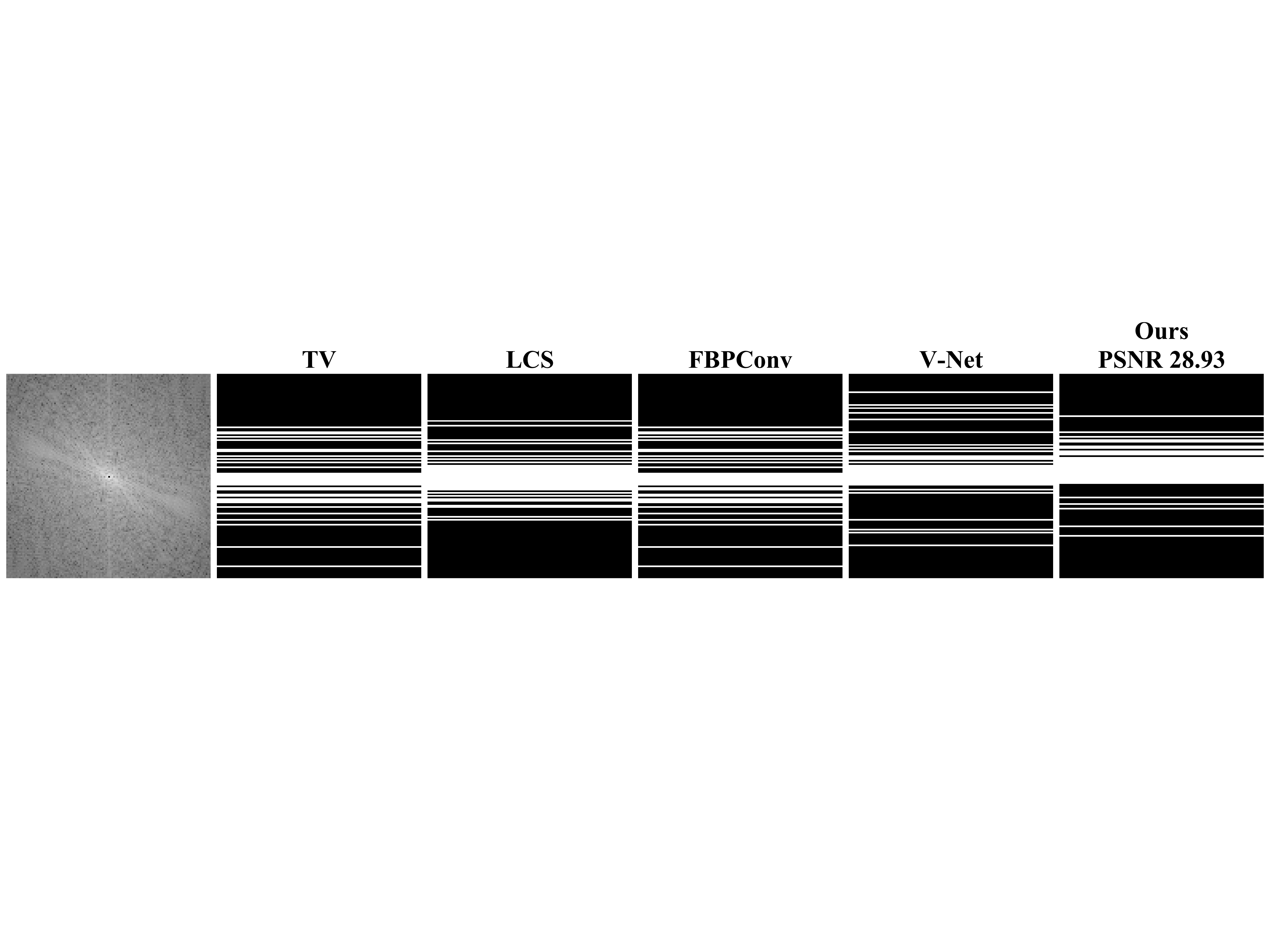}
\includegraphics[trim = 0mm 55mm 0mm 58mm,clip=true,width=17 cm]{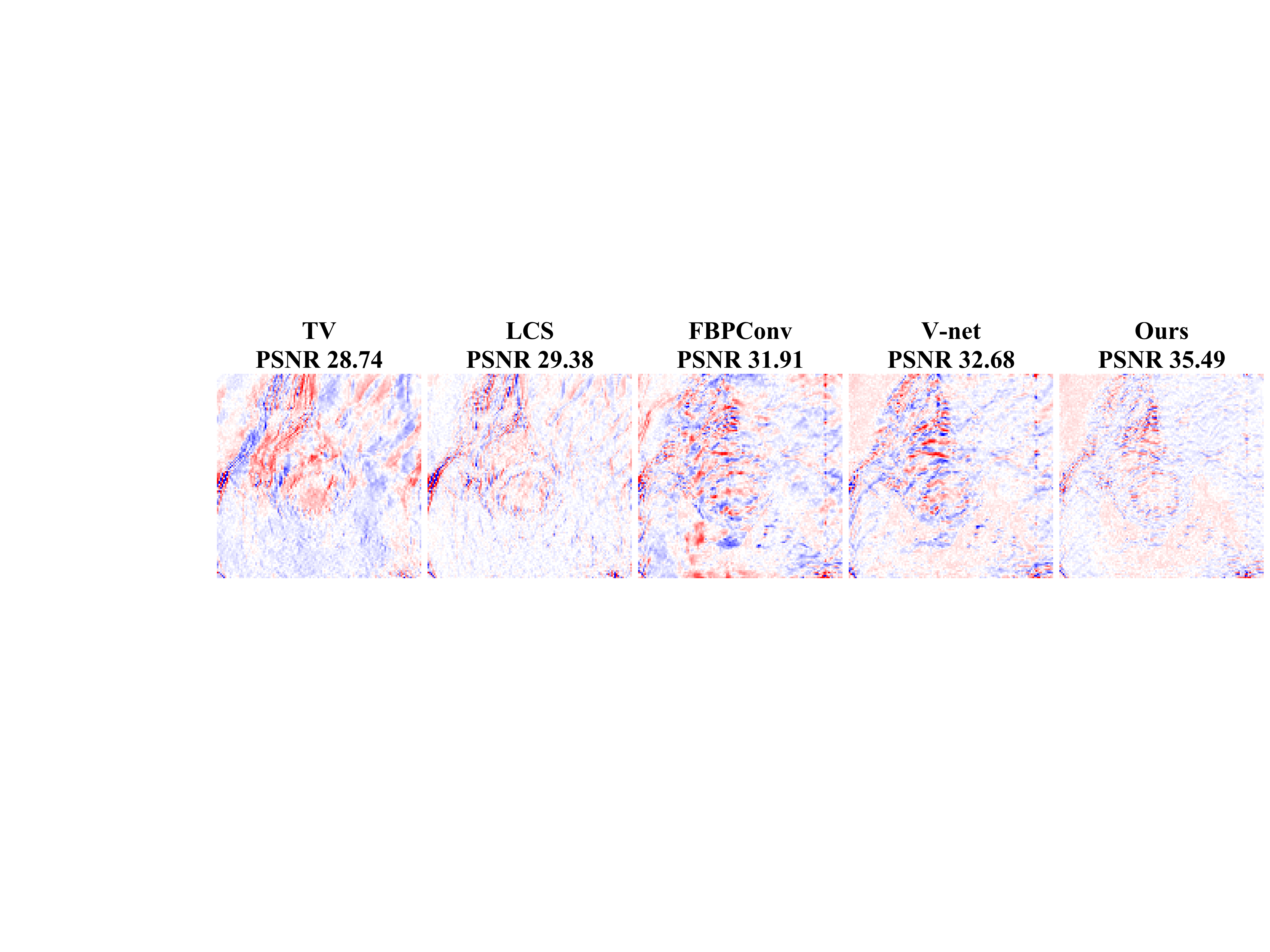}

\includegraphics[trim = 0mm 58mm 0mm 50mm,clip=true,width=17 cm]{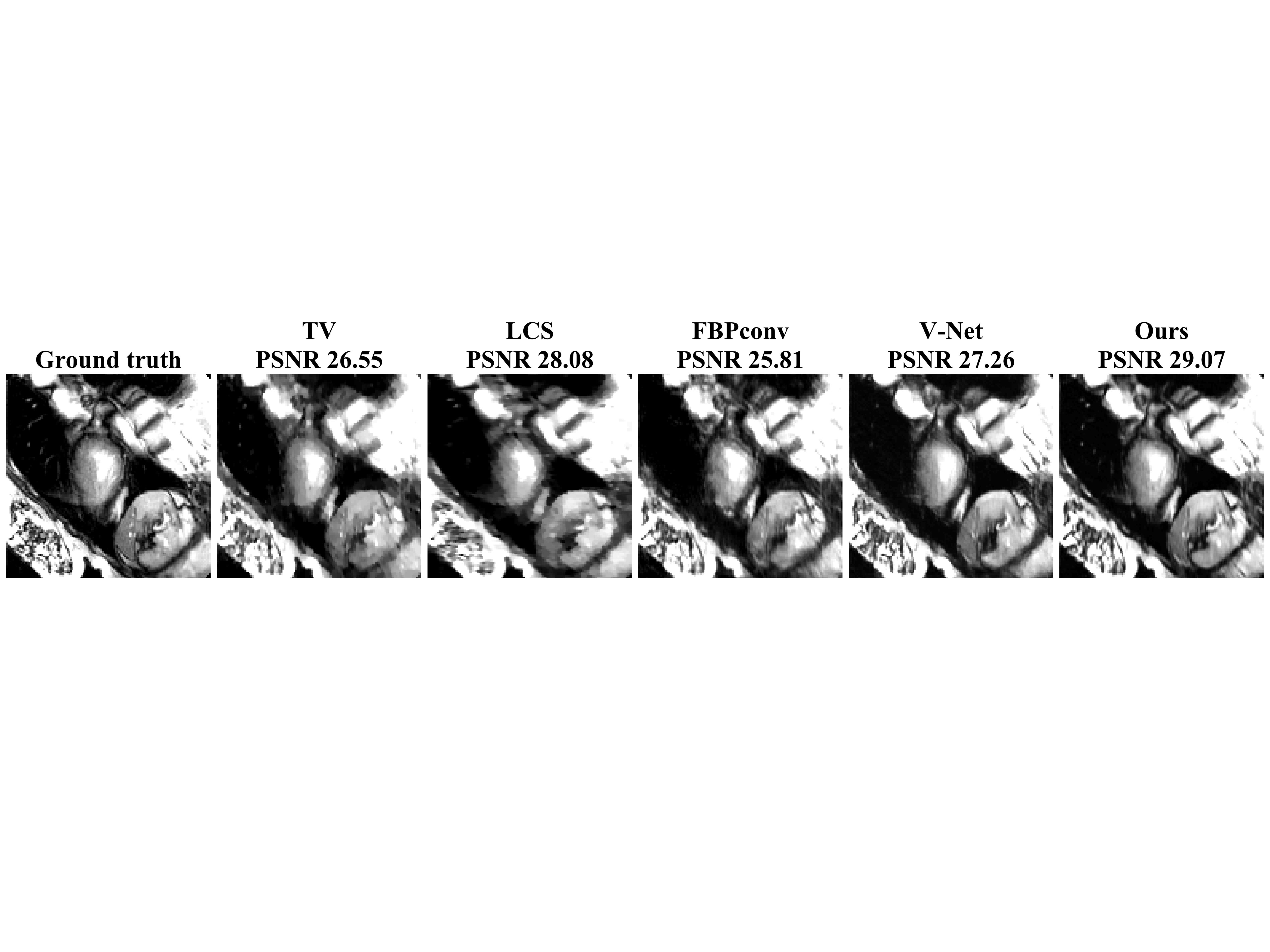}
\includegraphics[trim = 0mm 58mm 0mm 58mm,clip=true,width=17 cm]{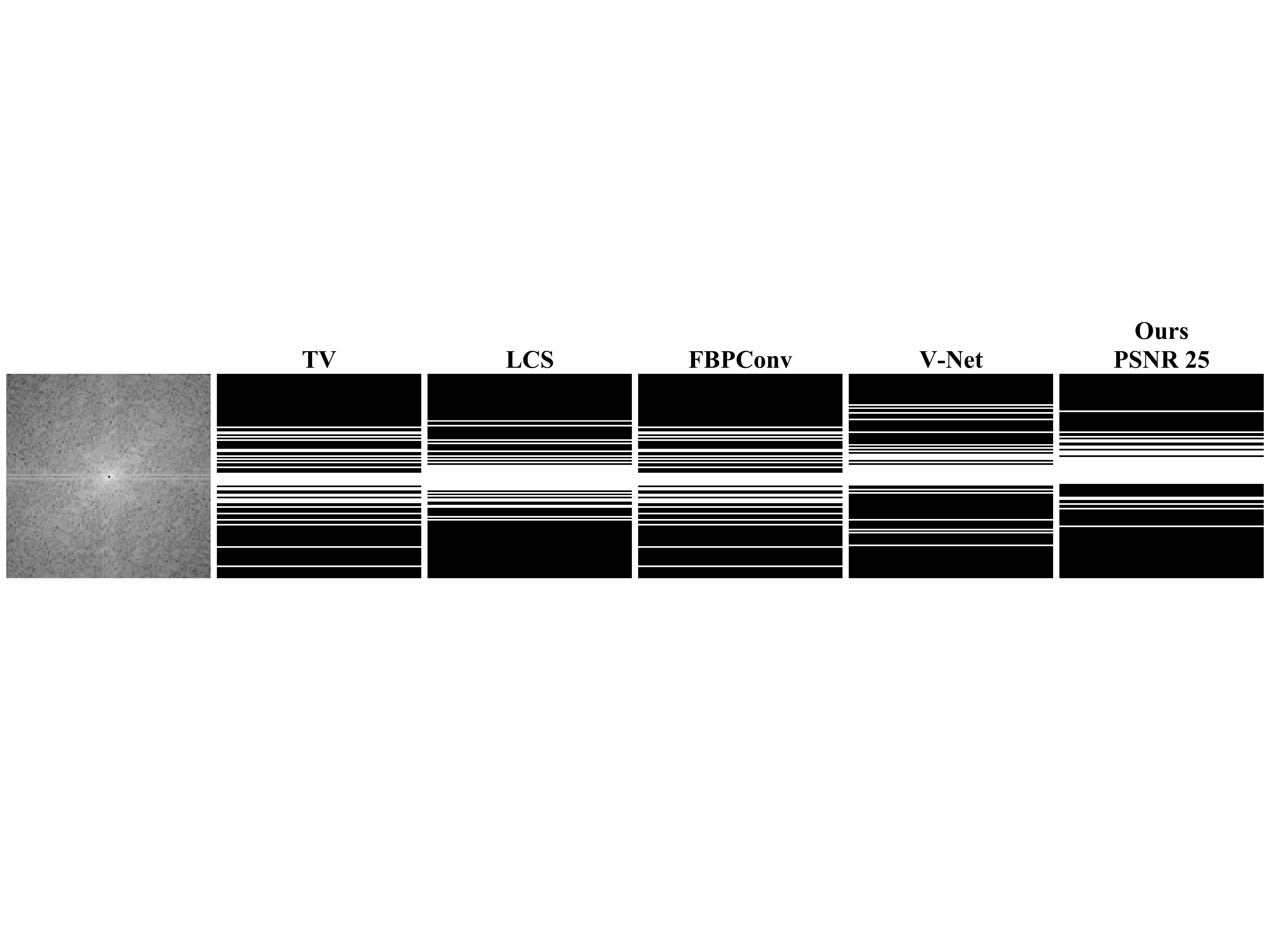}
\includegraphics[trim = 0mm 58mm 0mm 58mm,clip=true,width=17 cm]{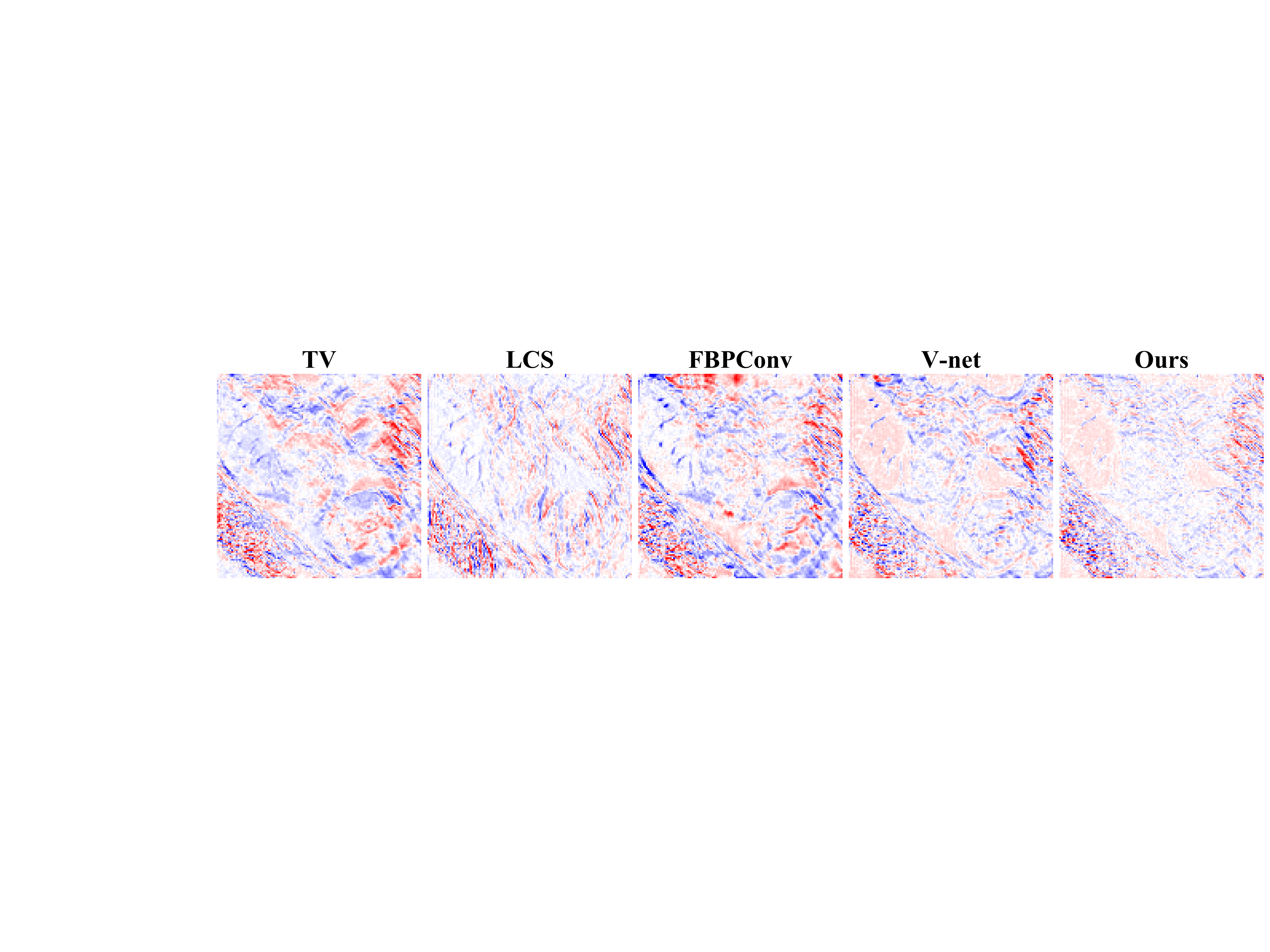}
\caption{Qualitative examples for the {\it cardiac} dataset.  (First row)
  reconstruction results; (second row) sampling patterns; (third row)
  color-coded residual between reconstruction and ground-truth, where red
  denotes positive and blue denotes negative errors.  See text for details.}
\label{fig:recon_c}
\end{figure*}


\begin{figure*}[t]
\centering
\includegraphics[trim = 0mm 58mm 0mm 50mm,clip=true,width=17 cm]{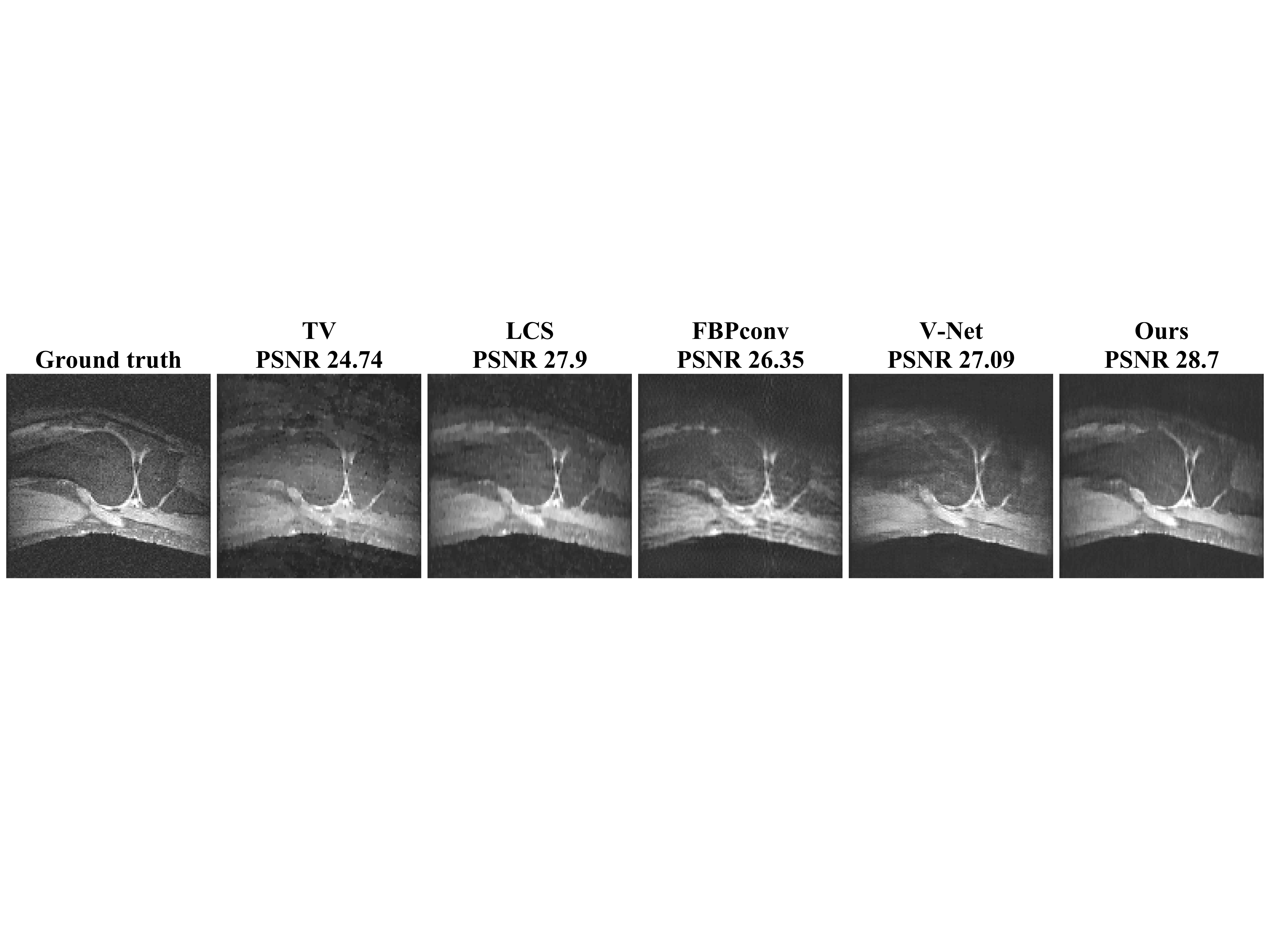}
\includegraphics[trim = 0mm 58mm 0mm 58mm,clip=true,width=17 cm]{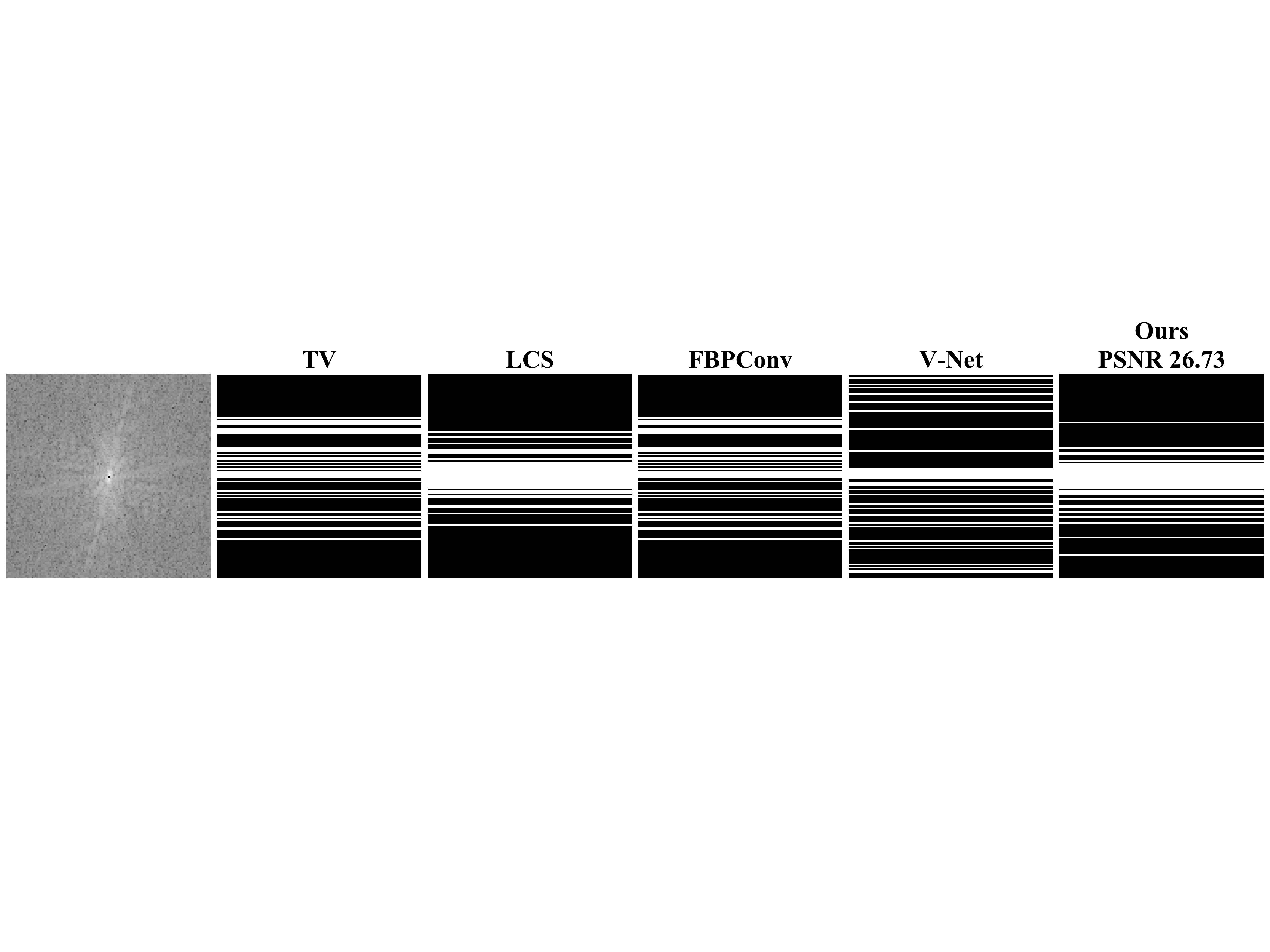}
\includegraphics[trim = 0mm 55mm 0mm 58mm,clip=true,width=17 cm]{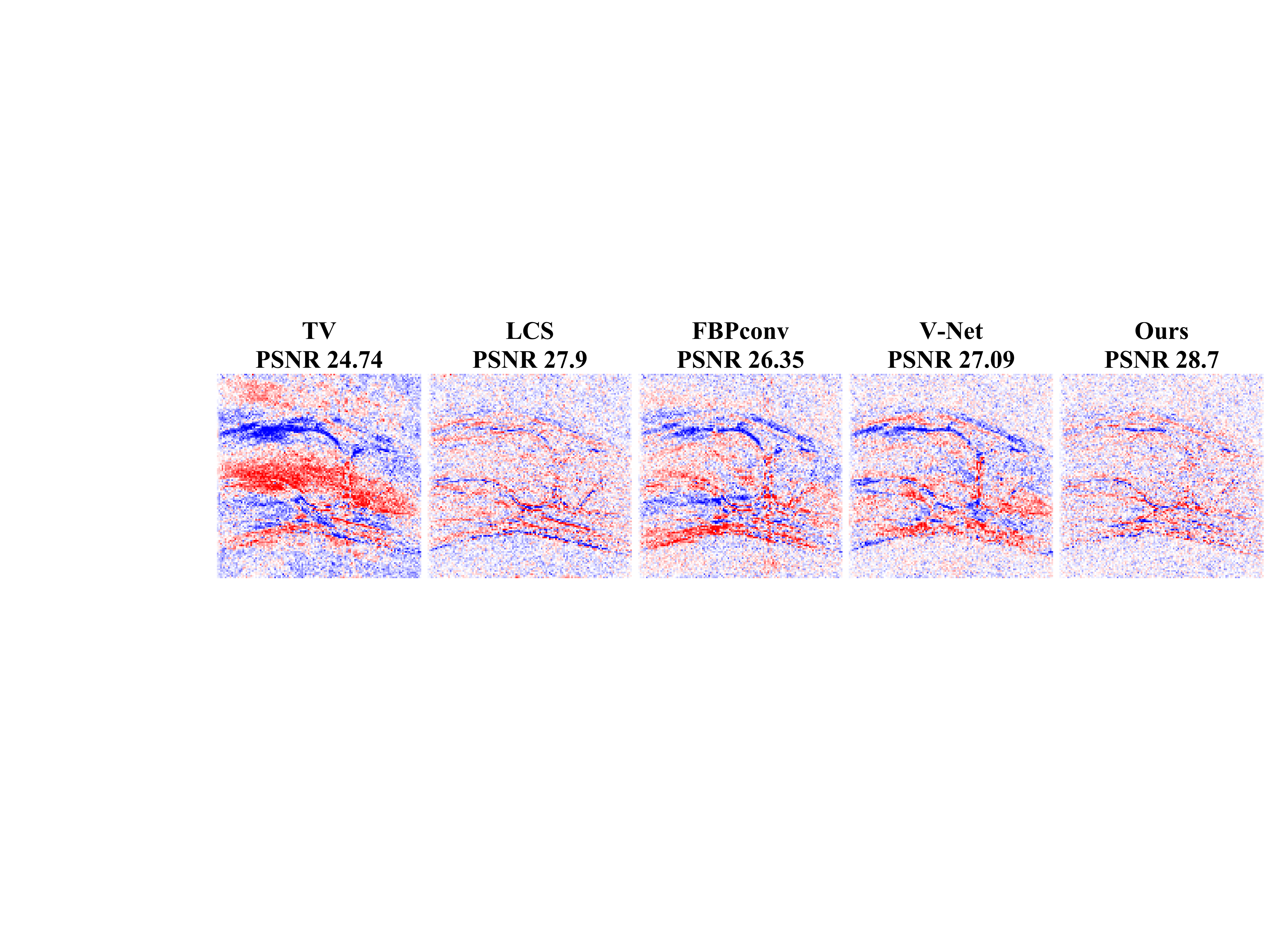}
\includegraphics[trim = 0mm 58mm 0mm 50mm,clip=true,width=17 cm]{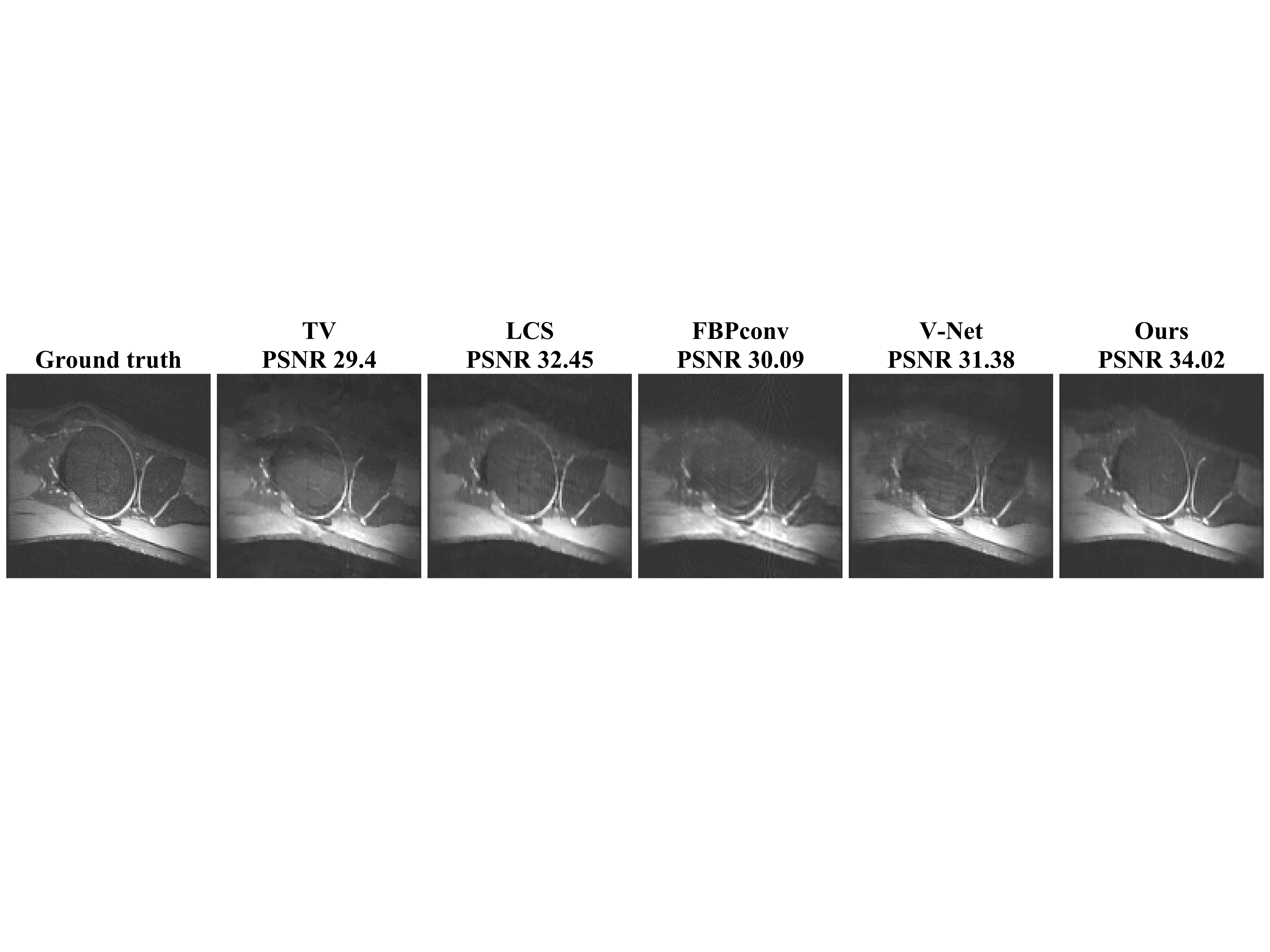}
\includegraphics[trim = 0mm 58mm 0mm 58mm,clip=true,width=17 cm]{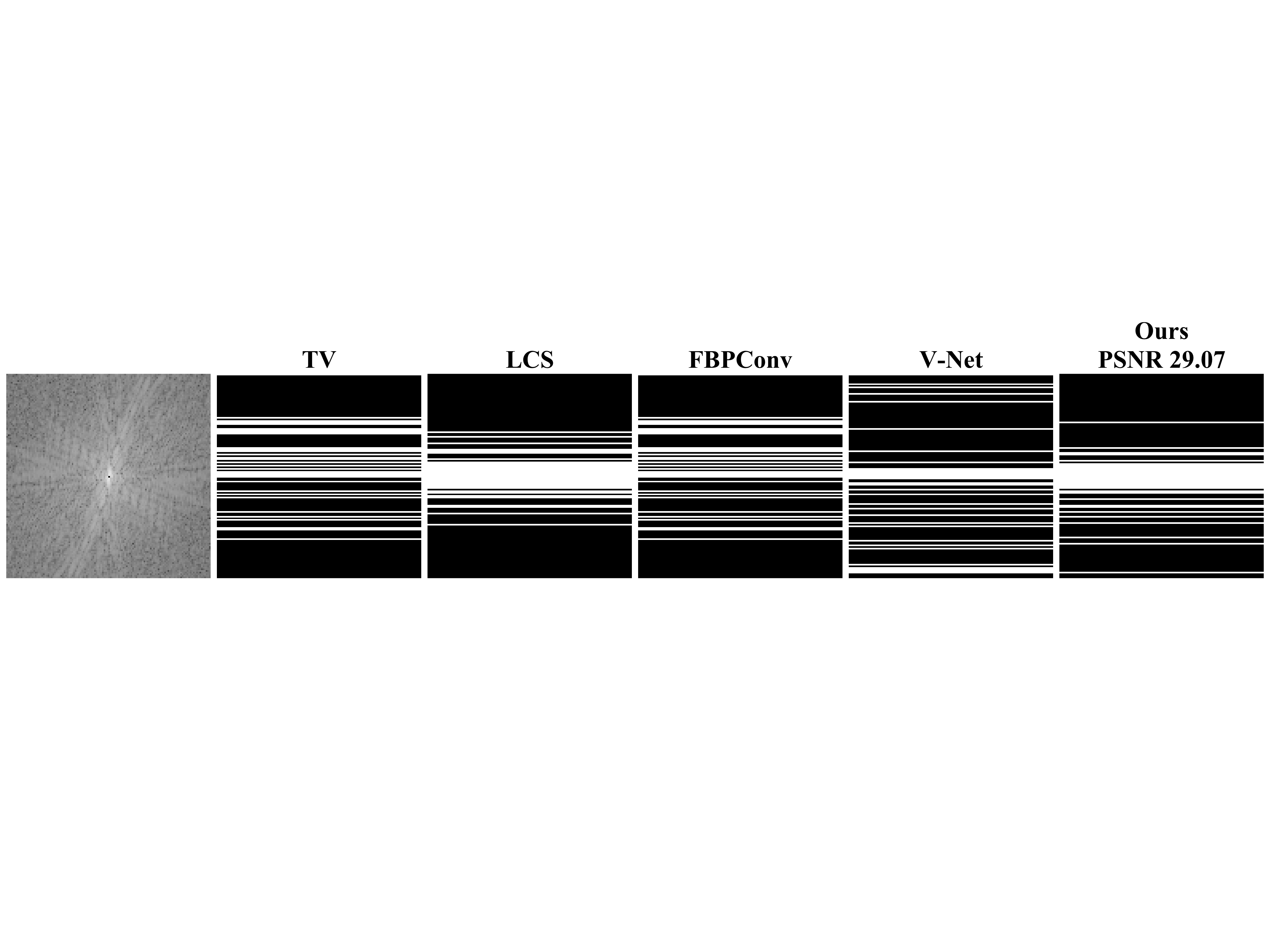}
\includegraphics[trim = 0mm 58mm 0mm 58mm,clip=true,width=17 cm]{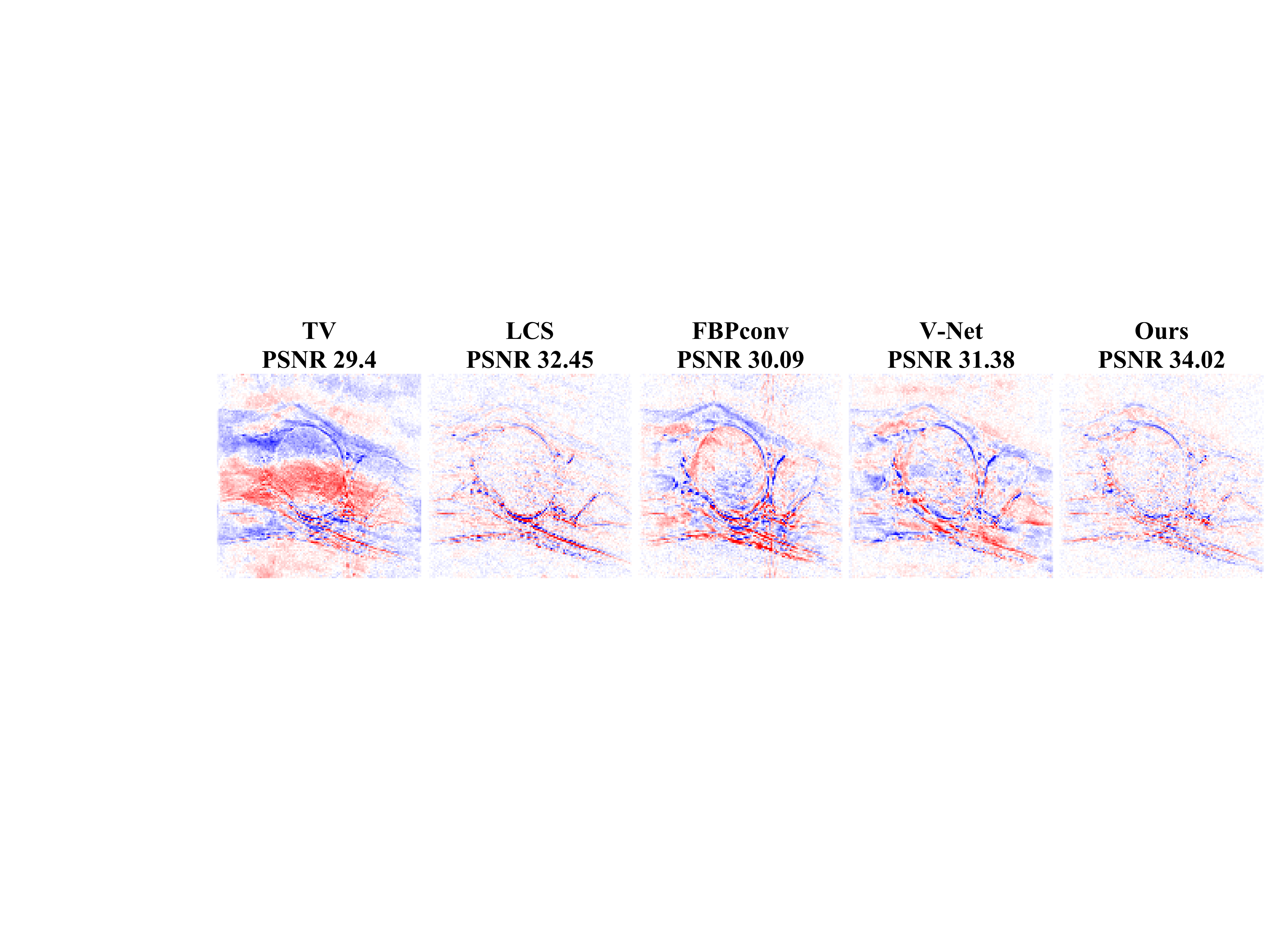}
\caption{Qualitative examples for the {\it knee} dataset.  (First
      row) reconstruction results; (second row) sampling patterns;
    (third row) color-coded residual between reconstruction and
    ground-truth, where red denotes positive and blue denotes negative
    errors.}
\label{fig:recon_k}
\end{figure*}


Qualitative examples for the {\it cardiac} dataset are shown in \fig{recon_c},
and for the {\it knee} dataset in \fig{recon_k}. In all examples, the proposed
method gives the highest reconstruction quality.

In \fig{recon_c}, as shown from the reconstruction examples of TV (VDS+TV) and
LCS (LCS+TV), cartoon-like textures are apparent. However, these artifacts are
well removed by our method. This is most apparent in \fig{recon_k}
bottom. Similarly, in TV and LCS of \fig{recon_k} (top), the textures of the
knee is blurry and appears as flat textures.  In FBPConv (VDS+FBPConv) and
V-Net (Ours+V-Net) , folded artifacts still remain. Our method is able to
clearly reconstruct the inner structures.

\subsection{t-SNE Analysis}

\begin{figure*}[t]
\centering
\includegraphics[trim = 50mm 10mm 50mm 10mm,clip=true,width=0.95\textwidth]{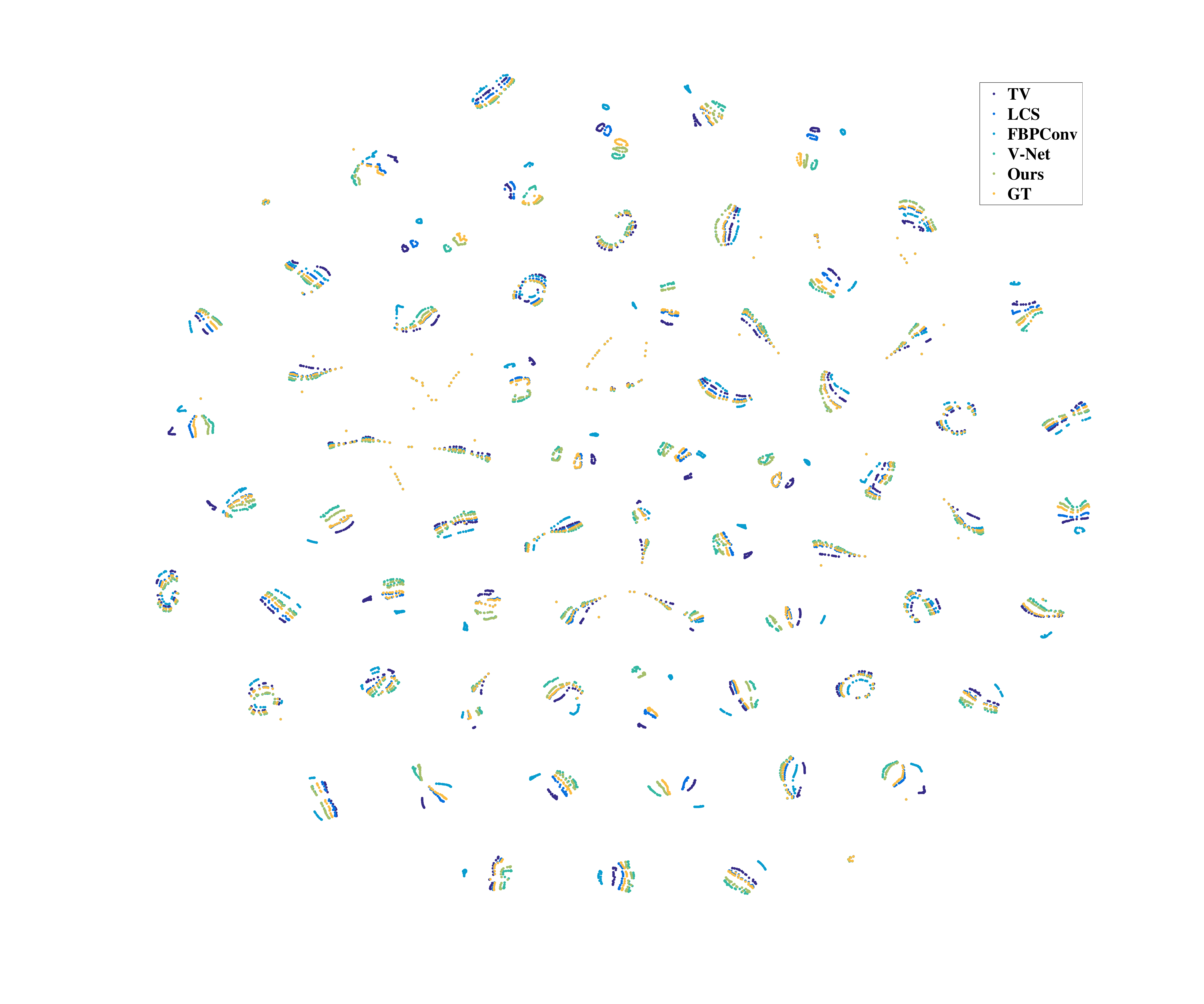}
\vspace{-10mm}
\caption{t-SNE \cite{maaten2008visualizing} embedding of the reconstructions
  for the testset of the {\it cardiac} dataset.}
\label{fig:tsne_c}
\end{figure*}


For biomedical applications, it is critical that the reconstructed data indeed
represents the original signal.  We therefore perform a t-SNE
analysis~\cite{maaten2008visualizing} to show that the reconstructed signals
indeed overlap the ground truth, distribution-wise. We use the {\it cardiac}
dataset for this analysis. It provides the largest corpus of test data for a
meaningful t-SNE analysis.

We show in \fig{tsne_c} the t-SNE map of the test set of the {\it cardiac}
dataset. Each point in the t-SNE map corresponds to an image reconstructed from
a different configuration. To avoid clutter, we only show the results of
one-fifth images, chosen randomly from a uniform distribution. As shown in
\fig{tsne_c}, the distribution and structure of our reconstructions are very
close to the ground truth.

\subsection{Sampling Evolution}

\begin{figure}
\centering
 \includegraphics[trim = 0mm 0mm 10mm 0mm,clip=true,width=0.95\linewidth]{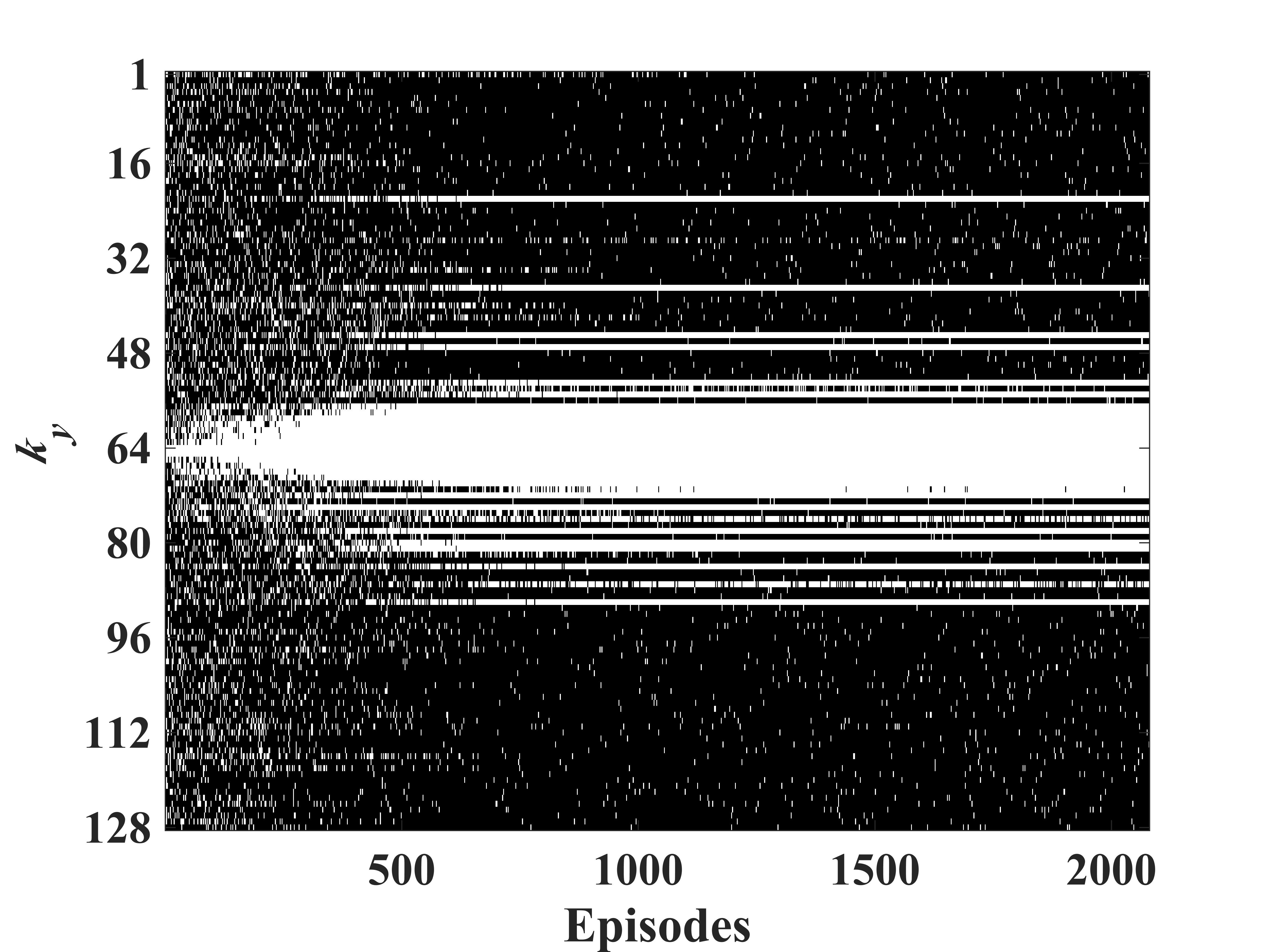}
 \caption{Evolution of the sampling patterns across optimization.  The vertical
   axis $k_y$ corresponds to sampling positions. White pixels denote that the
   position is sampled, whereas black ones denote it is not. The sampling
   pattern starts from random noise at the beginning, and converges to a main
   sampling pattern accompanied with minor changes depending on the image. }
\label{fig:evolution}
\end{figure}


In \fig{evolution}, we show how the final sampling pattern evolves as learning
proceeds. At first, it is essentially random, eventually converging to a main
sampling pattern that is shared for all training images plus minor components
that are image-dependent.  The fact that the main component is dominant is not
surprising, given that the MRI data have a similar appearance.

\subsection{Ablation Study}

We conduct ablation studies to motivate our design choices. We first show that
approximating the simulation phase with a deep network does not provide the
best performance. We then demonstrate the importance of implicit minimax backup
and, finally, the effectiveness of linking SampleNet with ReconNet.

\subsubsection{Importance of MCTS Rollout}

\begin{table}
  \caption{Quantitative results in terms of PSNR, with (V-Net) and without (our
    method) the deep network approximation for the reward in
    \cite{silver2017mastering}.  Best results are shown in bold.  }
    \label{tbl:vnet}
    \centering
    \begin{tabularx}{\linewidth}{C{1}|C{1}C{1}}
      \toprule
      Dataset &  With V-Net~\cite{silver2017mastering} & Our Method \\
      \midrule
      {\it cardiac} & 31.48 & {\bf 34.22} \\
      {\it knee} & 26.49 & {\bf 29.10} \\
      \bottomrule
    \end{tabularx}
  \end{table}


In \cite{silver2017mastering} it was suggested that using a value network,
referred to as the V-Net, that approximates the simulation outcomes can
speed-up the learning and lead to better results. We have also tried that
strategy. We provide in \tbl{vnet} the results compared to using a full
simulation, also dubbed roll-out. As the deployment of V-net gives
significantly worse results, we have not retained this strategy in our method.

\subsubsection{Effectiveness of Implicit Minimax Backup}

\begin{table}
  \caption{Quantitative results in terms of PSNR, with different MCTS $\alpha$
    values. Best results are shown in bold. }
    \label{tbl:mcts_im_alpha}
    \centering
    \begin{tabularx}{\linewidth}{C{1}|C{1}}
      \toprule
      minimax backup $\alpha$ & {\it  cardiac} \\
      \midrule
       $0$ & 32.99 \\
       $0.5$ &   {\bf 34.22} \\
       $1$ & 28.96 \\
      \bottomrule
    \end{tabularx}
  \end{table}


In \tbl{mcts_im_alpha}, we run our method with the backup parameter
$\alpha=0, 0.5, 1.0$. In principle, $\alpha$ lies between 0 and 1.  We observed
that our method performs best.
 
\subsubsection{Using ReconNet Output as SampleNet Input}

\begin{table}
  \caption{Quantitative results in terms of PSNR, with ReconNet (our method)
    and without ReconNet. Best results are shown in bold. }
    \label{tbl:worecon}
    \centering
    \begin{tabularx}{\linewidth}{C{1}|C{1}C{1}}
      \toprule
      Dataset &  Without ReconNet & Our Method \\
      \midrule
      cardiac & 33.53 & {\bf 34.22} \\
      \bottomrule
    \end{tabularx}
  \end{table}


In \tbl{worecon}, we compare two results. On one hand, we let the output of
ReconNet be the input the SampleNet. On the other hand, we bypass the ReconNet
process when training SampleNet. We use the {\it cardiac} dataset for this
experiment.  We see a faster\footnote{The time to convergence was four days.}
convergence rate when ReconNet is used, as well as better final results.  Thus,
we conclude that linking the two networks is helpful when learning.


\section{Conclusion}
\label{sec:conclusion}

We have proposed a self-supervised deep-learning framework that learns to
jointly sample and reconstruct accelerated MRI.  Our method is composed of two
networks, namely, SampleNet and ReconNet. They learn to sample and reconstruct
progressively. The two networks are trained with supervision signals that are
generated through Monte Carlo tree search.  This method provides on-the-fly
sampling patterns that improve the quality of reconstruction. ReconNet also
uses these sampling patterns to learn to reconstruct. By learning to do jointly
sampling and reconstruction, our framework is able to outperform the state of
the art.

To the best of our knowledge, our framework is the first to incorporate data
acquisition in the training process of deep networks in the context of MRI. We
believe this is a promising direction that goes beyond the current
deep-learning-based methods.  In the future, we hope to implement our method in
an actual MRI device, for high-speed data acquisition.


\section*{Acknowledgments}
This work was supported in part by the European Research Council under Grant 692726 (H2020-ERC Project GlobalBioIm), by the Natural Sciences and Engineering Research Council of Canada (NSERC) Discovery Grant “Deep Visual Geometry Machines” (RGPIN-2018-03788), and by systems supplied by Compute Canada.

\bibliographystyle{IEEEtran}
\bibliography{string,biomed,learning,optim,vision}

\end{document}